\pdfoutput=1
\PassOptionsToPackage{dvipsnames,svgnames,table}{xcolor}
\documentclass[11pt]{article}
\usepackage{authblk}
\usepackage[]{ACL2023}

\usepackage{hyperref}
\usepackage{booktabs}
\usepackage{graphicx}
\usepackage{enumitem}
\usepackage{xcolor}
\usepackage{multicol}
\usepackage{tabulary}
\usepackage{amsmath}
\usepackage{bold-extra}
\usepackage{subcaption}
\usepackage{float}
\usepackage{nicefrac}
\usepackage{color, colortbl}
\usepackage{times}
\usepackage{latexsym}
\usepackage{makecell}
\usepackage{amssymb} 
\usepackage{multirow}
\usepackage{tablefootnote}
\usepackage{siunitx}
\usepackage{textcomp}
\usepackage{amssymb}
\usepackage{pifont}
\newcommand{\cmark}{\ding{51}}%
\newcommand{\xmark}{\ding{55}}%

\usepackage{enumitem}
\renewcommand{\footnotesize}{\fontsize{8.5pt}{10.5pt}\selectfont}

\usepackage[T1]{fontenc}

\usepackage[utf8]{inputenc}

\usepackage{microtype}

\renewcommand\footnotemark{}

\newcommand{\mytilde}{\raise.17ex\hbox{$\scriptstyle\mathtt{\sim}$}}
\newcommand{\webie}{\textsc{WebIE}}

\title{\textsc{WebIE}: Faithful and Robust Information Extraction on the Web}

\author[1,2,*]{Chenxi Whitehouse\thanks{$^*$ Work conducted as Research Intern at Amazon Alexa AI.}}
\author[2]{Clara Vania}
\author[2,**]{Alham Fikri Aji\thanks{$**$ Now at Mohamed Bin Zayed University of Artificial Intelligence (MBZUAI), Abu Dhabi, UAE.}} 
\author[2]{\\Christos Christodoulopoulos} 
\author[2]{Andrea Pierleoni}
\affil[1]{City, University of London}
\affil[2]{Amazon Alexa AI, Cambridge, UK} 
\affil[ ]{\tt chenxi.whitehouse@city.ac.uk}
\affil[ ]{\tt \{vaniclar, chrchrs, apierleoni\}@amazon.co.uk}

\begin{document}
\maketitle
\begin{abstract}
Extracting structured and grounded fact triples from raw text is a fundamental task in Information Extraction (IE).
Existing IE datasets are typically collected from Wikipedia articles, using hyperlinks to link entities to the Wikidata knowledge base. 
However, models trained only on Wikipedia have limitations when applied to web domains, which often contain noisy text or text that does not have any factual information.
We present \textsc{WebIE}, the first large-scale, entity-linked closed IE dataset consisting of 1.6M sentences automatically collected from the English Common Crawl corpus.
\textsc{WebIE} also includes negative examples, i.e. sentences without fact triples, to better reflect the data on the web.
We annotate \mytilde21K triples from \textsc{WebIE} through crowdsourcing and introduce m\textsc{WebIE}, a translation of the annotated set in four other languages: French, Spanish, Portuguese, and Hindi.
We evaluate the in-domain, out-of-domain, and zero-shot cross-lingual performance of generative IE models and find models trained on \textsc{WebIE} show better generalisability. 
We also propose three training strategies that use entity linking as an auxiliary task. Our experiments show that adding Entity-Linking objectives improves the faithfulness of our generative IE models\footnote{Dataset, code and additional materials are available at \url{https://github.com/amazon-science/WebIE}.}.

\end{abstract}

\section{Introduction}

Information Extraction (IE) is the task of extracting structured information from unstructured text, usually in the form of triples \textit{<subject, relation, object>}. 
It is essential for many Natural Language Processing applications such as knowledge base population, question answering, faithful summarisation, and fake news detection \citep{trisedya-etal-2019-neural,huguet-cabot-navigli-2021-rebel-relation, narayan-etal-2021-planning, whitehouse2022evaluation}.

Typically, two pieces of information are needed for training closed IE\footnote{Closed IE refers to the extraction of triples with the entity and relation defined by a knowledge base.} systems: (i) the entities mentioned in the text and (ii) the relations that exist between each pair of entities. 
Obtaining such information requires expensive annotations, therefore most existing IE datasets, such as WikiNRE \citep{trisedya-etal-2019-neural} or REBEL \citep{huguet-cabot-navigli-2021-rebel-relation},
are built using Wikipedia, as entity information is available through hyperlinks and relation information can be automatically extracted via distant supervision (DS) approach \citep{mintz-etal-2009-distant} using a knowledge base (KB) such as Wikidata. 
The DS approach assumes that if two entities are connected through a relation in a KB, then the sentences that mention both entities together express the relation.

While models trained only on this fact-rich domain\footnote{We use the term \textit{domain} to refer to the URL domain.} have shown to be useful for IE applications, they have limited capacity when applied to extracting information in other web domains, which often contains noisy text or text without any factual information.
Take AllenAI's C4 dataset\footnote{We use the dataset from \url{https://huggingface.co/datasets/allenai/c4.}}, an open-sourced version of Google's C4 \cite{JMLR:v21:20-074} dataset based on Common Crawl, as an example.
Our analysis using the DS approach reveals that less than 15\% of the sentences contain triples (\S\ref{sec:auto}), whereas we observe that a state-of-the-art (SOTA) generative IE model, GenIE \cite{josifoski-etal-2022-genie}, which is trained on REBEL, the largest IE dataset to date (which includes only positive examples), tends to generate triples for every sentence, resulting in a high rate of false positives and issues with hallucination.

\begin{figure*}[t!]
\centering
    \includegraphics[width=\linewidth]{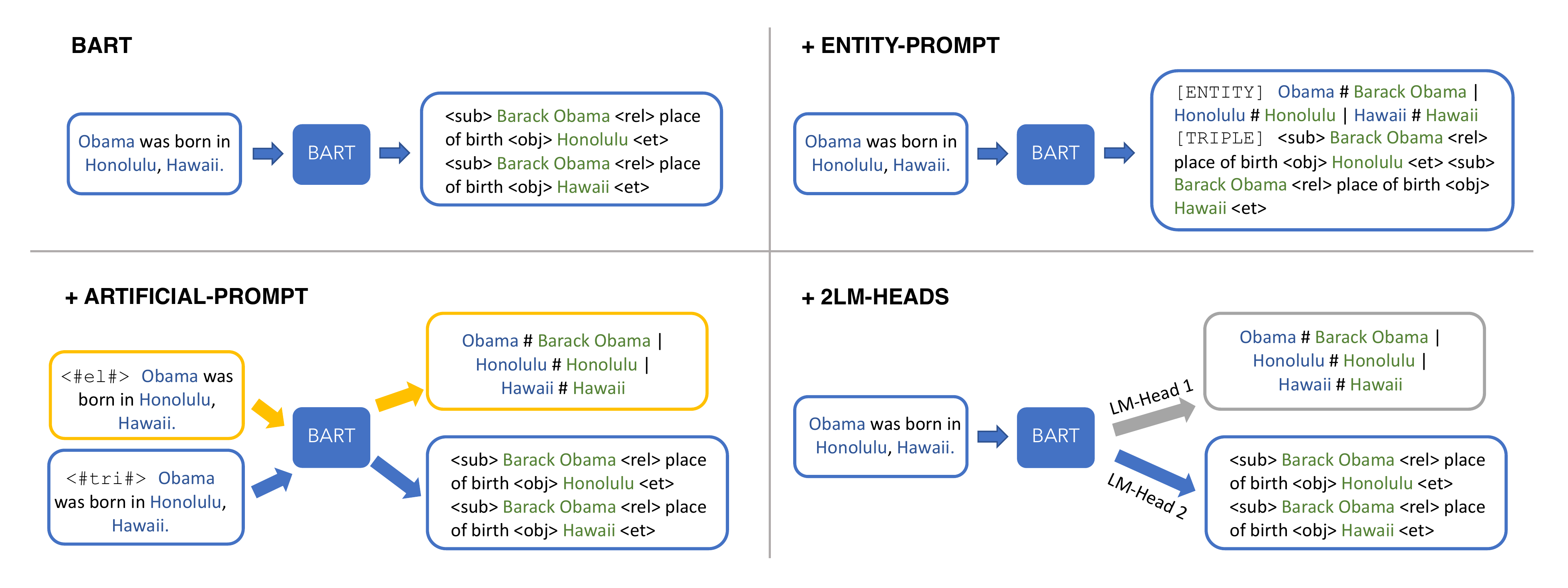}
    \caption{Training strategies used in this paper.
    The blue and green text refer to \textit{mention span} and its corresponding \textit{Wikipedia title} (used as entity labels).
    For standard BART training, the target output is the linearised triples (\S\ref{sec:linearised}).
    For \textsc{Entity-Prompt}, the target is the EL output (\S\ref{sec:joint}) concatenated with the linearised triples.
    In \textsc{Artificial-Prompt}, we prepend an artificial token to the input to indicate the desired output: EL (yellow box) or linearised triples. 
    For \textsc{2LM-Heads}, we add an additional task-specific LM head to the decoder for the EL task (grey box).}
\label{fig:webie_example}
\end{figure*}

To address these issues and facilitate future work on IE on the web, we present \textsc{WebIE}, the first large-scale, entity-linked closed IE dataset collected from web sources. 
The \textsc{WebIE} dataset is collected from the 200 most frequent URL domains from the C4 dataset. First, we use ReFinED \cite{ayoola-etal-2022-refined}, a state-of-the-art Entity Linking (EL) model to identify mention spans of the entities and link them to Wikidata.
We then apply the DS approach to extract triples and use a Natural Language Inference (NLI) model to filter out triples not expressed by the sentence.
We also include negative examples, i.e., sentences without any factual information, to better reflect the data on the web.
Our final dataset consists of 1.6M sentences, and we annotate a subset of \mytilde21K triples through crowdsourcing.
The annotated set is exclusively used as part of the test set to allow more reliable evaluation.
Finally, we introduce m\textsc{WebIE}, which contains human-corrected translations of the annotated version of \textsc{WebIE} in four languages: French, Spanish, Portuguese, and Hindi.

Previous works have shown that compared to discriminative pipelines which often suffer from accumulative errors due to separate Entity Linking and Relation Extraction (RE) steps \cite{mesquita-etal-2019-knowledgenet, trisedya-etal-2019-neural, josifoski-etal-2022-genie}, generative models achieve superior performance in many closed IE tasks.
Therefore we primarily benchmark \textsc{WebIE} with generative, transformer-based encoder-decoder models, BART \cite{lewis-etal-2020-bart} and mBART \cite{tang-etal-2021-multilingual}. 
The latter is used to evaluate the zero-shot cross-lingual transfer performance on m\textsc{WebIE}.

We further propose three training strategies (\S\ref{sec:joint}) that use entity linking as an auxiliary task for generative IE, namely joint generation with the linked-entity prompt (\textsc{Entity-Prompt}), multi-task learning with distinguished artificial prompt tokens (\textsc{Artificial-Prompt}), and training with an additional task-specific language model (LM) head (\textsc{2LM-Heads}).
We find that training with EL as an auxiliary task overall leads to better and more faithful IE results. 
An illustration of these training strategies is provided in \autoref{fig:webie_example}.

Our experiments show that compared to models trained only on Wikipedia datasets, models also trained on \textsc{WebIE} are more robust and generalisable, achieving a new SOTA performance on REBEL (\S\ref{sec:results}) and competitive zero-shot performance on WikiNRE.
We demonstrate that \textsc{WebIE} serves as a complementary dataset to existing datasets based on Wikipedia, and show that including negative examples is crucial for addressing false positives in generative IE.

Our main contributions are as follows:
(1) We present (m)\textsc{WebIE}, the first large-scale, entity-linked IE dataset on the web, where a subset is further annotated by humans and translated into four other languages;
(2) We propose and study the effectiveness of using entity linking as an auxiliary task for generative IE with various training strategies;
(3) Our comprehensive experiments demonstrate that models trained on \textsc{WebIE} exhibit better generalisability in Information Extraction on the web domain, including competitive zero-shot performance on IE tasks on Wikipedia.

\begin{table*}[!t]
\centering
\scalebox{0.7}{
\addtolength{\tabcolsep}{-1pt}
\begin{tabular}{ p{1.2cm}|p{1.8cm}p{1.1cm}p{1.2cm}p{1.5cm}p{1.4cm}p{1.4cm}p{1.4cm}p{1.5cm}p{1.5cm}p{1.5cm}p{1.5cm}}
\toprule
\multicolumn{1}{c|}{\sc{DATASET}} &{Domains} &  {Entity Linked} &Relation Types  & \multicolumn{1}{r}{Sentences} & \multicolumn{1}{c}{Train\textsuperscript{$\dagger$}} & \multicolumn{1}{c}{Validation\textsuperscript{$\dagger$}} &\multicolumn{1}{c}{Test\textsuperscript{$\dagger$}} & \multicolumn{1}{c}{Triples} &
{Annotated Triples}   &

Negative Instances
&
Languages (Test Set)\\

\midrule
\sc TACRED & Web
&\multicolumn{1}{c}{\xmark}
& \multicolumn{1}{r}{42} 
& \multicolumn{1}{r}{106,264}
& \multicolumn{1}{r}{68,124}
& \multicolumn{1}{r}{22,631}
& \multicolumn{1}{r}{15,509}
&\multicolumn{1}{r}{106,264}
&  \multicolumn{1}{r}{106,264}
&  \multicolumn{1}{c}{79.5\%} 
&  \multicolumn{1}{c}{1}\\
\sc WebRED & Web (120\textsuperscript{$\ddagger$})
& \multicolumn{1}{c}{\xmark}
& \multicolumn{1}{r}{523}
& \multicolumn{1}{r}{117,717}
& \multicolumn{1}{c}{--}
& \multicolumn{1}{c}{--}
& \multicolumn{1}{c}{--}
& \multicolumn{1}{r}{117,717}
&  \multicolumn{1}{r}{117,717} 
& \multicolumn{1}{c}{65\%}
&  \multicolumn{1}{c}{1}\\
\midrule
\sc WikiNRE & Wikipedia
&\multicolumn{1}{c}{\cmark}
& \multicolumn{1}{r}{158} 
& \multicolumn{1}{r}{255,654}
& \multicolumn{1}{r}{224,881}
& \multicolumn{1}{r}{988}
& \multicolumn{1}{r}{29,785}
& \multicolumn{1}{r}{330,005}
&  \multicolumn{1}{r}{0}
&  \multicolumn{1}{c}{0} 
&  \multicolumn{1}{c}{1}\\
\sc REBEL & Wikipedia 
&\multicolumn{1}{c}{\cmark}
& \multicolumn{1}{r}{1146}
&\multicolumn{1}{r}{3,059,894}
&  \multicolumn{1}{r}{2,754,387} 
& \multicolumn{1}{r}{152,672} 
& \multicolumn{1}{r}{152,835}  
& \multicolumn{1}{r}{10,311,293}
&  \multicolumn{1}{r}{0}
&  \multicolumn{1}{c}{0}
&  \multicolumn{1}{c}{1}\\
\sc WebIE & Web (200\textsuperscript{$\ddagger$}) 
& \multicolumn{1}{c}{\cmark} & 
\multicolumn{1}{r}{661}
& \multicolumn{1}{r}{1,649,167}
& \multicolumn{1}{r}{1,480,223}
& \multicolumn{1}{r}{82,914}
& \multicolumn{1}{r}{86,030}
& \multicolumn{1}{r}{1,905,205}
& \multicolumn{1}{r}{21,113}
& \multicolumn{1}{c}{50\%}
&  \multicolumn{1}{c}{5}\\
\bottomrule
\end{tabular}
}
\caption{Statistics of \textsc{WebIE} and comparison with other sentence-level RE (top two rows) and IE datasets. 
We report the publicly available version of WebRED.
$\dagger$ shows the number of examples in each split. 
$\ddagger$ corresponds to the number of URL domains.
Annotated Triples show the number of human-annotated triples.
}
 \label{tab:WebIE}
\end{table*}

\section{(m)\textsc{WebIE}}

In this section, we provide a detailed explanation of the dataset collection process for (m)\textsc{WebIE}.

\subsection{Collecting \textsc{WebIE}}
\label{sec:auto}
\paragraph{Data Preprocessing} 
We start with the English portion of the AllenAI's C4 dataset and keep the most frequent 200 URL domains\footnote{See \autoref{sec:domains} for URL domains included in \textsc{WebIE}.}. 
We randomly sample 1M documents and use SpaCy\footnote{\url{https://spacy.io/}} for sentence segmentation. Sentences with fewer than 10 words are removed, resulting in \mytilde20M sentences.

\paragraph{Entity Linking and DS Dataset} 
Next, we run ReFinED \citep{ayoola-etal-2022-refined}, a state-of-the-art EL model on the sentences to identify entity spans and link them to their corresponding Wikidata ID. 
Besides named entities, ReFinED also extracts \textit{numerical} entities that do not have Wikidata ID.
In this work, we only consider numerical entities that express dates, and map them to the corresponding \textit{year} for simplicity\footnote{For example, ``October 10, 2018'' will be mapped to ``2018''.}.
Some examples of ReFinED processed output are included in \autoref{sec:ReFinED}.

After obtaining the entity-linked sentences, we apply the DS paradigm to retrieve the set of relations that exist between each pair of entities in each sentence using Wikidata (September 2022 dump) as our KB and build a DS dataset.
After the above steps, we obtain \webie\ DS dataset consisting of 21.2M entities and 4.8M triples.

\paragraph{Entailment Filtering} 
One major drawback of the DS approach is that the triples extracted may or may not be expressed by the source sentence \citep{riedel2010}. Following previous work on obtaining a cleaner version of the DS dataset \citep{huguet-cabot-navigli-2021-rebel-relation,vania-lee-and-andrea-pierleoni-2022-improving}, we apply an NLI model, \texttt{nli-deberta-v3-large}\footnote{\url{https://huggingface.co/cross-encoder/nli-deberta-v3-large} achieved superior results among the models we evaluated in our preliminary experiments.}, that is trained on SNLI \cite{bowman-etal-2015-large} and MultiNLI \cite{williams-etal-2018-broad},
to filter out triples that do not entail the sentence.

Each source sentence is treated as the \textit{premise} and we use manually created templates (similar to \citet{vania-lee-and-andrea-pierleoni-2022-improving}) to convert a DS triple to one or more \textit{hypotheses}.
%
We then obtain the entailment probability score for each premise-hypothesis pair and take the maximum score for cases with multiple converted hypotheses.
We set the threshold to be 0.7, similar to \citet{huguet-cabot-navigli-2021-rebel-relation}, and only keep triples with an entailment score above the threshold. We retain 2.1M triples (44\% of the previous DS triples, see \autoref{tab:WebIE}) after this filtering process.

\paragraph{Negative Examples} 
After the DS creation and NLI filtering steps, only less than 10\% of the original sentences contain triples.  
To train models for extracting facts from the web and alleviate false positives, we include two kinds of negative examples in \textsc{WebIE}:
(i) sentences with one or zero entities, and 
(ii) sentences with two or more entities, but without any factual information (i.e., no relation between the entities). 
We randomly sample negative instances covering both cases evenly and add them to \textsc{WebIE}.
In the end, \textsc{WebIE} consists of 1.6M sentences, where 50\% are negative examples.
A summary of the statistics of \textsc{WebIE} with a comparison with other datasets is shown in \autoref{tab:WebIE}.
The dataset is randomly split into train/validation/test sets using a 90/5/5 split.

\subsection{Human Annotation}
Existing IE datasets, such as REBEL, are often automatically annotated using the DS approach, hence the labels can be noisy.
To allow more reliable evaluation of \webie, we randomly sample \mytilde21K triples from the most frequent 200 relations and annotate them with MTurk.
Given a sentence, each HIT (Human Intelligence Task) is designed to verify if a DS triple is correctly expressed in the sentence\footnote{We ensure \textit{all} DS triples in a selected sentence are annotated.}.
First, the annotators are asked to verify if the head entity (subject) and tail entity (object) are linked correctly.
For each entity, we provide its Wikipedia title and link to its Wikidata page as additional context. 
After that, the annotators are asked to verify if the triple relation is correctly inferred from the sentence. Here, we provide the relation descriptions and example use cases of each relation.
We ask three MTurk workers to annotate each DS triple and take the majority vote as the final label for each triple.
A triple is considered valid if both entities are linked to the correct Wikidata entities and the relation is inferred\footnote{We ask for \textit{inferred} instead of explicit expression since some relations may not be explicitly expressed in the sentence, e.g. ``located in'' (London, UK) or ``date of birth''  XX (1986-2022).} by the sentence.
An annotation interface is shown in \autoref{sec:mturk}.

To ensure the annotation quality, we set qualifications with additional requirements for MTurk workers (see \autoref{sec:mturk} for details).
The agreement among the three annotators is high: 99.4\% for the head entities, 99.2\% for the tail entities, and 76.1\% for the relations have all three annotators agreeing on the same label.
After the majority vote, 92.1\% of the triples are labelled as inferred and therefore kept as valid triples.

\subsection{Multilingual \textsc{WebIE}}
\label{sec:mwebie}
To enable zero-shot cross-lingual transfer evaluation on \webie, we further extend the annotated subset, with additional negative examples, to four other languages: French, Spanish, Portuguese, and Hindi.
First, we use a neural machine translation model, the distilled 1.3B variant\footnote{\url{https://huggingface.co/facebook/nllb-200-distilled-1.3B}} of NLLB-200 \cite{costa2022no}, to translate the English sentences into the target languages.
We then use MTurk to verify the translation and add entity span information in the translated sentences. 
We provide the English sentence (with the entity spans highlighted) and its translation, and first, ask the annotators to correct the translation. After that, MTurk workers are asked to mark the corresponding entity spans in the target language.
We ask two annotators to complete the aforementioned HIT, and an additional worker to select the better translation, which is used in our final dataset.
To obtain translations with higher quality, we restrict the region of the workers to countries where the target language is the official language\footnote{See details for m\textsc{WebIE} annotations in \autoref{sec:mturk}.}.
The final m\webie\ consists of 9K instances in each language, which corresponds to roughly 90\% of the 21K annotated triples.

\section{Generative Information Extraction}

This section describes the training strategies that we use for benchmarking (m)\textsc{WebIE}. 

\subsection{Sentence-to-Triples Generation}
\label{sec:linearised}
We use BART and mBART for all of our experiments.
Given a sentence $s$ as input, we train the model to autoregressively generate the linearised  triples $t$ as an output.
Following the practice from \citet{huguet-cabot-navigli-2021-rebel-relation} and  \citet{josifoski-etal-2022-genie}, we linearise a triple $t_i$ by converting it into ``\textit{<sub> head entity label <rel> relation <obj> tail entity label <et>}'', where the tags in brackets represent \textbf{sub}ject, \textbf{rel}ation, \textbf{obj}ect, and the \textbf{e}nd of \textbf{t}riple, respectively. 
Head/tail entity label refers to the Wikipedia title that the mention span in the sentence is mapped to, which also has a one-to-one correspondence with the Wikidata ID\footnote{For example, a mention span of ``UK'' is linked to Wikipedia title ``United Kingdom'' and mapped to Q145 in Wikidata.}.

For each sentence, we order its linearised triples accordingly to the order in which they appear in the input sentence; first by the order of the appearance of the head entity, and then by the order of the tail entity (for cases when the head entities are the same).
The conditional probability of generating $t$ is formulated as $p(t|s) = \prod_{t=0}^{N} p(t_i |t_{<i}, s)$. 
We use the standard cross-entropy loss and maximise the output sequence likelihood with teacher forcing \cite{NIPS2014_a14ac55a}.
An example of input and output can be seen in the top left of \autoref{fig:webie_example}.

\subsection{Entity-Linking as an Auxiliary Task}
\label{sec:joint}

The standard linearised triples output only contains the label of the entity and not the span. As a result, it may be difficult to trace back from which input span an entity is generated, especially in the case when the model hallucinates (e.g., by generating an entity that is not mentioned in the sentence).
To encourage models to generate faithful and interpretable output, we also experiment with models that are jointly optimised for generating triples and EL.
The goal of the EL task is to identify and extract entity spans from the input sentence and link them to their corresponding KB entities.
We posit that adding the EL task as an additional training objective will teach the model to put attention to the input spans when generating the output.
We experiment with the following three approaches.

\paragraph{\textsc{Entity-Prompt}}
\citet{narayan-etal-2021-planning, narayan-etal-2022-well} have shown that generation with entity-chain planning, i.e. generating the desired entities first before the actual output, is effective in improving the faithfulness and controlling hallucinations in text generation tasks such as abstractive summarisation.
For generative IE tasks, EL can be used as an intermediate plan to ground the generation of the linearised triples.
We define the Entity-Linking target in the format of ``\textit{Mention Span\textsubscript{1} \# Entity Label\textsubscript{1} | Mention Span\textsubscript{2} \# Entity Label\textsubscript{2} | ...}'', where the entity spans are ordered as they appear in the text.
We then prepend the EL target to the linearised triples target, using special symbols as separators, i.e., ``\textit{[ENTITY] Entity-Linking target [TRIPLE] Linearised Triples Target}'', where ``\textit{[ENTITY]}'' is the start symbol before generating the EL output, and ``\textit{[TRIPLE]}'' is the start symbol before generating the linearised triples. 
Given an input sentence, we essentially train the decoder to first generate the EL chain and then generate the triples, conditioned on both the input sentence and the EL output\footnote{The EL target only includes mention spans that contribute to valid triples, consistent with the triples that are later generated conditioned on the linked entities.}.

\paragraph{\textsc{Artificial-Prompt}} 
Artificial Prompt tokens are symbols placed in front of the input sequence, which has previously been explored in areas such as neural machine translation to distinguish the language of the target output translation \cite{johnson-etal-2017-googles}, visual question answering for joint answer and explanation generation \cite{whitehouse-etal-2023-towards}, etc.
We adapt this approach for jointly training our models for Entity Linking and generative IE.
Specifically, we use an artificial prompt token \texttt{<\#el\#>} at the beginning of the input sentence when training for the Entity-Linking target, and use \texttt{<\#tri\#>}\footnote{Both artificial prompt tokens are added as the special tokens to the tokenizer to avoid bias from pre-trained embeddings, but are intended to be biased to the associated task.} for the linearised output target.
Training instances for both tasks are mixed and randomly shuffled for training.

\paragraph{\textsc{2LM-Heads}}
Finally, inspired by \citet{gontier2022does}, the third approach that we experiment with is the addition of a second language model (LM) head in the decoder, which is initialised with the same weights as the first (standard) LM head.
The first LM head is optimised for generating the linearised triples while the second LM head is optimised for the EL task, thus each training instance has two different target outputs.
During training, the input sentence is fed to the encoder once, and different target outputs are given to the \textit{same} decoder. Each task-specific LM head is then responsible for generating output targeted for it.
The training loss is then formulated as a weighted sum of the losses from both tasks: $\mathcal{L}=\alpha \mathcal{L}\textsubscript{IE} + (1-\alpha) \mathcal{L}\textsubscript{EL}$.

\begin{table*}[!ht]
\centering
\scalebox{0.7}{
\addtolength{\tabcolsep}{-3pt}
\begin{tabular}{l|cccrcccrcccccccccc}
\toprule
{\multirow{2}{*}{\sc Model}}  &\multicolumn{4}{c} {\sc WebIE (all test)}   &\multicolumn{4}{c}  {\sc WebIE (anno. test)}& \multicolumn{3}{c} {\sc REBEL}   & \multicolumn{3}{c} {\sc Wiki-NRE}
\\
& \textit{Precision} &\textit{ Recall} & \textit{F1}  & \textit{Acc.-Neg.}& \textit{Precision} &\textit{ Recall} & \textit{F1}  & \textit{Acc.-Neg.} & \textit{Precision} &\textit{ Recall} & \textit{F1 }& \textit{Precision} & \textit{Recall} &\textit{ F1 }  \\
\cmidrule(lr){1-1} 
\cmidrule(lr){2-5} \cmidrule(lr){6-9} \cmidrule(lr){10-12}  \cmidrule(lr){13-15} 

\sc Bart\textsubscript{rand} (r) 		
&  \cellcolor[HTML]{DDEBF7}11.93
&  \cellcolor[HTML]{DDEBF7}18.91
& \cellcolor[HTML]{DDEBF7}14.63 
& \cellcolor[HTML]{DDEBF7}0.00
&  \cellcolor[HTML]{DDEBF7}11.82		
&  \cellcolor[HTML]{DDEBF7}15.63
& \cellcolor[HTML]{DDEBF7}13.46  
& \cellcolor[HTML]{DDEBF7}0.00
& 66.89&70.37&	68.58 
&\cellcolor[HTML]{DDEBF7}27.61
&\cellcolor[HTML]{DDEBF7}66.73
&\cellcolor[HTML]{DDEBF7}39.06\\
\sc Bart\textsubscript{plm} (r) 		
&  \cellcolor[HTML]{DDEBF7}15.24
&  \cellcolor[HTML]{DDEBF7}39.30 
& \cellcolor[HTML]{DDEBF7}21.96
&  \cellcolor[HTML]{DDEBF7}0.00
&  \cellcolor[HTML]{DDEBF7}15.98		
&  \cellcolor[HTML]{DDEBF7}34.92
& \cellcolor[HTML]{DDEBF7}21.93 
& \cellcolor[HTML]{DDEBF7}0.00
&66.28&	76.78 &	71.14
&\cellcolor[HTML]{DDEBF7}25.39
&	\cellcolor[HTML]{DDEBF7}77.45
&	\cellcolor[HTML]{DDEBF7}38.24
\\
\cmidrule(lr){1-15}
\sc Bart\textsubscript{rand} (w)
&  55.47 &	57.25&	56.35&90.07
&  52.95&	46.60&	49.57&95.04
&\cellcolor[HTML]{DDEBF7}27.47
&\cellcolor[HTML]{DDEBF7}23.13
&	\cellcolor[HTML]{DDEBF7}25.12
&\cellcolor[HTML]{DDEBF7}18.98
&	\cellcolor[HTML]{DDEBF7}43.75
&\cellcolor[HTML]{DDEBF7}26.48
\\
\sc Bart\textsubscript{plm} (w) 
& 57.92 & 74.19 & 64.91 & 87.99
& 57.00 & 65.91 & 61.13 & 94.18
&\cellcolor[HTML]{DDEBF7}35.81 
&\cellcolor[HTML]{DDEBF7}43.00 
&\cellcolor[HTML]{DDEBF7}39.08
&\cellcolor[HTML]{DDEBF7}24.30 
&\cellcolor[HTML]{DDEBF7}78.01 
&\cellcolor[HTML]{DDEBF7}37.06

\\
\cmidrule(lr){1-15}
\sc Bart\textsubscript{rand} (r+w)  
&52.79 &64.15 &57.92 & 87.45
&51.89& 54.28 & 53.06  &  93.71
&66.87 &72.24 &69.45
&\cellcolor[HTML]{DDEBF7}29.02
& \cellcolor[HTML]{DDEBF7}82.35
&\cellcolor[HTML]{DDEBF7}42.91
\\
\sc Bart\textsubscript{plm} (r+w)
&54.63&78.43&64.40 & 76.43
&55.22&71.25&62.22&82.59
&66.42 &78.29 & 71.87
& \cellcolor[HTML]{DDEBF7}29.25 
& \cellcolor[HTML]{DDEBF7}86.38
& \cellcolor[HTML]{DDEBF7}43.70 
\\

\bottomrule
\end{tabular}
}
\caption{Experiment results with constraint Trie.
\textsc{Bart\textsubscript{rand}} corresponds to models with BART configuration but randomly initialised weights.
\textsc{Bart\textsubscript{plm}} are models with pretrained weights from \citet{lewis-etal-2020-bart}.
\textsc{(r)}, \textsc{(w)}, \textsc{(r+w)} refer to models trained on REBEL, \textsc{WebIE}, and both datasets, respectively. 
For \webie\ we show the overall performance and the accuracy on negative samples.
The blue shade indicates zero-shot performance.
}
\label{tab:main-results}
\end{table*}

\begin{table*}[!ht]
\centering
\scalebox{0.75}{
\addtolength{\tabcolsep}{-2pt}
\begin{tabular}{l|rrrrc|rrrrc}
\toprule
{\multirow{2}{*}{\sc Language}}
 &\multicolumn{5}{c|} {\sc Unconstrained Decoder}  &\multicolumn{5}{c} {\sc Constraint Trie}
\\

&\textit{Precision} &\textit{ Recall} & \textit{F1} & \textit{Empty-Pos.\%} 
& \textit{Accuracy-Neg.} 
&\textit{Precision} &\textit{ Recall} & \textit{F1} & \textit{Empty-Pos.\%}  & \textit{Accuracy-Neg.} 

\\
\cmidrule(lr){1-1} 
\cmidrule(lr){2-4} \cmidrule(lr){5-5} \cmidrule(lr){6-6} \cmidrule(lr){7-9}  
\cmidrule(lr){10-10} 
\cmidrule(lr){11-11} 
\sc English 
&57.72 &61.26 &59.43 &2.48 &95.69
&60.29 &64.29 & 62.22 &2.63 &96.11
\\
 \sc French 
&43.27 &36.13 &39.38 &11.89 &96.19
&46.52 & 40.26 & 43.16 &12.63 &96.64
\\
 \sc Spanish  
&41.93 & 34.63 & 37.93 &12.34 & 96.74
&45.13 &38.89 & 41.78 & 12.80 &96.97
\\
\sc Portuguese 
&41.17 &32.37 &36.24 &14.07 &96.91
&44.15 & 36.61 & 40.02 & 14.82 &97.22
\\

 \sc Hindi 
 &4.28 & 1.62 & 2.35 &67.38& 98.64
&4.23 & 1.67 & 2.40 & 67.55 & 98.64
\\
\bottomrule
\end{tabular}
}
\caption{Performance on m\textsc{WebIE} with mBART. Results for non-English are zero-shot. 
Empty-Pos(itive)\% shows \textit{false} negatives, revealing zero-shot performance has a high rate of empty results for positive examples.
}
\label{tab:positive-negative-mwebie}
\end{table*}

\subsection{Inference with a Constraint Trie}
In addition to standard beam search decoding, we experiment with constraint decoding by restricting the generated output to be valid Wikipedia titles and Wikidata relations using a prefix Trie, following the ideas proposed in GENRE \cite{cao2021autoregressive} and GenIE \cite{josifoski-etal-2022-genie}. 
%
We use two constraint Tries: an entity Trie and a relation Trie. The entity Trie is built using all Wikipedia titles (as the entity labels),
and the relation Trie is built using all Wikidata relation property labels. 
We refer the readers to \citet{cao2021autoregressive} for more details on constructing the Trie. 

We use four special symbols, \textit{<sub>}, \textit{<rel>}, \textit{<obj>} and \textit{<et>} to define the state of the generation. 
We apply both constraint Tries as follows. 
We adopt the constraint Trie so that, in the very first decoding state, the model is allowed to either (i) return an empty string for a negative example, or (ii) generate \textit{<sub>}, which is the start symbol for generating a triple.
If the \textit{<sub>} symbol is generated, then we generate the head entity using the entity Trie, i.e., only valid entities will be considered.
Once the generation of the head entity is completed, the model proceeds to generate \textit{<rel>} (i.e., the start symbol for generating relation string) and then subsequently generate allowed tokens from the relation Trie which is built from the relations in Wikidata.
After that, the model generates \textit{<obj>} and the tail entity, in the same manner, using the entity Trie.
After generating the full triple (indicated by \textit{<et>} generated after the tail entity), the decoder can either stop the generation or start a new iteration for generating the next triple.

For the \textsc{Entity-Prompt} models, since the entity mention spans are text from the input sentences and usually are not the same as the entity labels in Wikidata, we propose a \textit{partial} constraint generation approach. 
Specifically, we start the standard beam search for the EL target output and only activate the Trie constraints after that when generating the linearised triples.


\section{Experiments}

In this section, we explain the datasets used in the experiments and the detailed modelling setup.

\subsection{Dataset}
\label{sec:dataset}
In addition to our proposed \textsc{WebIE} dataset, we also use the following datasets for our experiments.

\paragraph{WikiNRE}\citep{trisedya-etal-2019-neural} is an IE dataset based on Wikipedia which is automatically constructed by aligning Wikipedia sentences to Wikidata triples using the DS approach.
The authors apply a coreference resolution model \cite{clark-manning-2016-improving} to obtain sentences with implicit entity names, and use a paraphrase detection model \cite{ganitkevitch-etal-2013-ppdb,grycner-weikum-2016-poly} to filter out sentences that do not express the DS triples. In our experiments, we only use WikiNRE for zero-shot evaluation.

\paragraph{REBEL} \citep{huguet-cabot-navigli-2021-rebel-relation} is a large-scale IE dataset constructed automatically from Wikipedia abstracts. 
Using the Wikipedia hyperlinks in the abstracts, as well as numerical values and dates, they map the entity spans to their corresponding Wikidata entities. They then use the DS approach to identify triples in each sentence.
To filter out false positives, the authors use an NLI model by \textit{concatenating} the entities and the relation as the hypothesis.
In our experiment, we use the REBEL dataset that is sub-sampled by \citet{josifoski-etal-2022-genie}, where 857 relations are considered. Both WikiNRE and REBEL do not contain negative examples and are not annotated by humans.

\begin{table*}[!ht]
\centering
\scalebox{0.75}{
\addtolength{\tabcolsep}{-1.5pt}
\begin{tabular}{l|cccccc|cccccc}
\toprule
{\multirow{3}{*}{\sc Model}} & \multicolumn{6}{c|} {REBEL} & \multicolumn{6}{c} {\sc {WebIE (anno)}} \\
& \multicolumn{3}{c} {\sc Unconstrained} & \multicolumn{3}{c|} {\sc Constraint Trie} & \multicolumn{3}{c} {\sc Unconstrained} & \multicolumn{3}{c} {\sc Constraint Trie} \\
 & \textit{Precision} & \textit{Recall} & \textit{F1 }& \textit{Precision} &\textit{ Recall} & \textit{F1} & \textit{Precision} &\textit{ Recall} & \textit{F1 }& \textit{Precision} & \textit{Recall} &\textit{ F1 }
\\
\cmidrule(lr){1-1} 
\cmidrule(lr){2-4} \cmidrule(lr){5-7} \cmidrule(lr){8-10} \cmidrule(lr){11-13}

 \sc Bart\textsubscript{rand}
 & 64.34	&67.90	& 66.07 
 & 66.89	&70.37&	68.58
 & 51.64 &	44.46 &	47.78 	
&  52.95 &	46.60 &	49.57\\
\sc Entity-prompt 
& 63.30	& 63.04	& 63.17	
& 67.91 & 	67.54 &	67.72 
& 49.64 & 51.62 & \textbf{50.61}
& 51.90 & 54.28 & \textbf{53.06}
\\
\sc Artificial-prompt 
&64.23 &	68.23&	66.17 
& 66.41&	70.72&	68.50 
& 52.33 & 46.21 & 49.08
& 53.86 & 48.18 & 50.86
\\
 \sc 2lm-heads & 65.16 &	68.70  &	\textbf{66.88}
 & 67.05&	70.88&	\textbf{68.91}
& 49.13 & 47.67 & 48.39 
& 51.07 & 49.59 & 50.32
 \\
\bottomrule
\end{tabular}
}
\caption{Comparison of various training with entity linking as an auxiliary task, and beam search with and without constraint Trie decoding. 
\textsc{WebIE} results are on the annotated test set.
All models use BART configuration with randomly initialised weights. 
We show in bold the best F1 scores among the training objectives.
}
\label{tab:el-models} 
\end{table*}

\subsection{Models}
We experiment with BART using two settings: \textsc{Bart\textsubscript{plm}} with the pre-trained weights from \citet{lewis-etal-2020-bart}\footnote{\url{https://huggingface.co/facebook/bart-large}}, and \textsc{Bart\textsubscript{rand}}, using the same configuration and architecture but randomly initialised weights.
Across the two settings, \citet{josifoski-etal-2022-genie} find that \textsc{Bart\textsubscript{rand}} generates better results than \textsc{Bart\textsubscript{plm}} on REBEL.
For m\textsc{WebIE}, we experiment with the mBART-50\footnote{\url{https://huggingface.co/facebook/mbart-large-50}} model (for simplicity we refer to it as mBART in this paper).

To compare models trained on different datasets, we train both \textsc{Bart\textsubscript{plm}} and \textsc{Bart\textsubscript{rand}} on REBEL \textsc{(r)}, \textsc{WebIE} \textsc{(w)}, and both datasets together \textsc{(r+w)}.
We evaluate the performance of the generated triples by parsing the linearised output to a list of triples and comparing it to the gold label to calculate precision, recall, and F1 scores.
For \textsc{WebIE}, we also calculate the accuracy of the prediction of negative instances, where a prediction is considered correct if the model accurately generates empty strings rather than hallucinating triples.

For training with EL as an auxiliary task, we primarily experiment with the \textsc{Bart\textsubscript{rand}}. 
We prepare the training instances as described in \S\ref{sec:joint}, and train separate models on REBEL and on \textsc{WebIE}.
For the \textsc{2LM-Heads}, we conduct experiments with different values of the $\alpha$ parameter in the combined loss function, specifically, we set it to 0.5 and 0.75.

We use 8 GPUs, each with 32G VRAM, for all experiments.
We set the batch size to 8 and accumulate gradient batches to 32.
We follow the hyperparameters settings from \citet{josifoski-etal-2022-genie} and set the learning rate to $3e^{-5}$, weight decay to 0.01, and warmup steps to 5K\footnote{For \textsc{Bart\textsubscript{plm}(w)} we find it is necessary to use a lower learning rate $5e^{-6}$ for more stable training.}.
We train for up to 30 epochs with early stopping (patience 10), validate twice per epoch, and take the last checkpoint for evaluation. 
Training one epoch takes \mytilde1.5 hours for BART and \mytilde2 hours for mBART.

\section{Results and Analysis}
\label{sec:results}
We now present the main results of (m)\textsc{WebIE} and compare different training strategies.

\subsection{Main Results}
\label{sec:main-results}
\autoref{tab:main-results} shows our benchmarking results on \textsc{WebIE}.
We report results with the constraint Trie in decoding since it overall achieves better results\footnote{See \autoref{tab:main-full} in \autoref{sec:main-full} for detailed comparison.}.
Contrary to the findings from \citet{josifoski-etal-2022-genie}, we find that BART models with pre-trained weights are better than initialised weights.
Constraint Trie decoding benefits REBEL, WikiNRE, and the recall performance of \textsc{WebIE}, but may compromise the precision since the models are also trained to handle negative examples.

Models trained on both REBEL and \textsc{WebIE} \textsc{(r+w)} obtain overall better F1 scores on the two datasets compared to models trained on each dataset separately.
Similar performance can also be observed in the zero-shot performance on WikiNRE.
Models trained solely on the REBEL dataset (Wikipedia-domain) show poor generalisability on \webie\footnote{For positive examples it only achieves 20 F1 points.} and always generate false positives thus resulting in 0\% accuracy for negative instances in \webie.
This indicates that Wikipedia-domain data only is not adequate for training robust models for the web, and the absence of negative examples in these datasets leads to a prominent issue of hallucination when applied to the web.

\textsc{Bart\textsubscript{plm} (r+w)} also achieves a new state-of-the-art F1 score of 71.87 on REBEL, surpassing the performance of 68.93 from GenIE \cite{josifoski-etal-2022-genie} and 70.74 from KnowGL \cite{knowgl-aaai_2023_demo}, the latter of which trains with additional information including entity type.
The results demonstrate the benefit of \textsc{WebIE}, which contributes to the generalisability of the models.

\subsection{Cross-lingual Transfer with mBART}

We train mBART on the training set of \webie\ and evaluate the zero-shot cross-lingual transfer on m\webie. 
Similar to prior experiments, results in \autoref{tab:positive-negative-mwebie} show that constraint Trie decoding obtains higher performance than standard decoding\footnote{We report results using EN as the source language token for mBART, as it produces better performance compared to the actual source language token. See more details in \autoref{sec:main-full}.}.

For English, mBART achieves higher overall performance than \textsc{Bart\textsubscript{plm}} (see \autoref{tab:main-results}). 
The zero-shot results reveal that Hindi has a significant decline in performance compared to the other three non-English languages, French, Spanish, and Portuguese. 
Since these three languages utilise the Latin script as in English, which may result in an overlap of entity surface forms. 
In contrast, the transfer is more difficult for Hindi as it employs a different writing system. 
Manual analysis indicates that mBART tends to produce a high rate of false negatives in Hindi examples, where the correct extraction mostly occurs when the entities in the sentences share similar surface forms with the English counterparts.


\subsection{Results with Additional EL Training}
\label{sec:aux}
\autoref{tab:el-models} shows the results of training with Entity-Linking as an auxiliary task.
For REBEL, the best results are achieved with the \textsc{2LM-Heads} approach, where the $\alpha$ parameter is set to 0.75.
For \textsc{WebIE} with negative examples, all EL training models achieve better F1 performance than \textsc{Bart\textsubscript{rand}}, with \textsc{Entity-Prompt} particularly resulting in better recall.
This shows the benefit of joint training with EL to improve the faithfulness of web domain data.
\textsc{Artificial-Prompt} achieves the best precision in \textsc{WebIE} but does not show significant differences in performance compared to \textsc{Bart\textsubscript{rand}}. 
Nevertheless, all three approaches provide better interpretability, i.e., the information of the mention spans in the text that contributes to the IE prediction.

\textsc{Entity-Prompt} and \textsc{Artificial-Prompt} do not require additional architectural adaptation over the standard model.
\textsc{Entity-Prompt} also does not introduce training overhead, whereas the other two models may require twice the training time.
\textsc{2LM-Heads} offers the flexibility of adapting the weighted combination of the main task and the auxiliary task by adjusting $\alpha$ in the joint loss formula, which allows more emphasis on the main target. 

\section{Related Work}

\paragraph{IE Datasets}
The term Information Extraction has been used for different tasks in the literature.
Most existing IE datasets are collected from Wikipedia articles aligned with Wikidata, including sentence-level IE datasets such as REBEL, WikiNRE, FewRel \cite{han-etal-2018-fewrel}, T-REx \cite{elsahar-etal-2018-rex}; document-level Relation Extraction\footnote{We consider RE dataset as the ones that focus on extracting relations but without entity spans and/or linking information.} datasets, e.g., DocRED \cite{yao-etal-2019-docred}, CodRED \cite{yao-etal-2021-codred}.   
SMiLER \cite{seganti-etal-2021-multilingual} is a multilingual sentence-level IE dataset that is also based on Wikipedia, covering 14 languages and 36 relations. 
These sentence-level IE datasets typically do not contain negative examples.

Datasets such as TACRED \cite{zhang-etal-2017-position}, RE-TACRED \cite{Stoica_Platanios_Poczos_2021}, and WebRED \citep{ormandi2021webred} have negative relation examples but they are not linked to knowledge bases. 
Our proposed dataset \textsc{WebIE} is distinct from the existing datasets in that it is on the web domain, entity-linked, and with negative examples.

\paragraph{IE Approaches}
IE approaches can be classified into two categories: pipeline systems with discriminative models, and sequence-to-sequence systems with generative models. 
Pipeline models typically include separate modules for Named Entity Recognition (NER), Entity Linking and Relation Extraction \cite{chaganty-etal-2017-importance, yamada-etal-2020-luke}.
Systems that jointly train NER, EL, and RE, have also been explored, taking advantage of the information shared among the tasks \cite{ji-etal-2020-span, span2020}.

In recent years, generative IE has gained a lot of attention.
\citet{Nayak_Ng_2020} utilise an LTSM model and propose a pointer network-based decoding.
More recent approaches, e.g. as introduced in REBEL and GenIE, train a transformer-based encoder-decoder model with standard maximum-likelihood objectives to convert sentences to linearised output.
KnowGL \cite{knowgl-aaai_2023_demo} improves upon REBEL with additional entity type information added to the linearised output.
Our work extends GenIE and experiments with three different approaches where we incorporate explicit EL information as an auxiliary task with adapted constraint Trie decoding.

\section{Conclusions}
We present (m)\textsc{WebIE}, the first large-scale, entity-linked closed IE dataset on the web.
A subset of the dataset is further annotated by humans and translated into four other languages, French, Spanish, Portuguese, and Hindi, via crowdsourcing.

We benchmark \webie\ with generative models and compare the models trained on \webie\ and REBEL (Wikipedia-domain).
Our results show that models trained on \webie\ have competitive zero-shot performance when applied to REBEL and WikiNRE, whereas models trained only on REBEL have 0\% accuracy on the negative examples in \webie. This highlights the importance of including negative examples for training more robust models and reducing hallucination in generative IE on the web. 
Models trained on both REBEL and \webie\ achieve the best performance on both datasets, as well as zero-shot results on WikiNRE, showing that \textsc{WebIE} serves as a complementary dataset to existing Wikipedia-domain datasets.

Investigating the approaches with Entity Linking as an auxiliary task, we find that adding an additional task-specific LM head achieves the overall best performance for REBEL, and the \textsc{Entity-Prompt} approach shows the most significant improvement on \webie, particularly benefiting recall.
We primarily benchmark transformer-based encoder-decoder models on \webie, but future work could also explore pipeline frameworks and larger language models for few-shot performance.


\section*{Limitations}
We identify several limitations in this work:
(i) \textbf{False negatives}. Our current automatic triple extraction pipeline is built using the DS approach followed by filtering using an NLI model.
However, Wikidata is not complete \citep{tan-etal-2022-revisiting}.
While some triples may not be completely available in \webie, we expect models trained on this dataset can still discover new triples that do not exist in Wikidata.
(ii) \textbf{Limited relations in annotation}. The human annotation is only conducted on the most frequent 200 relations.
(iii) \textbf{Limited languages in m\textsc{WebIE}}. As discussed in \S\ref{sec:mwebie} and \autoref{sec:mturk}, the languages in m\textsc{WebIE} are limited to official languages from geographical regions where there is a reasonable amount of MTurk workers to accept the job. 
An alternative solution would be to use professional translators, especially for low-resource languages.
(iv) \textbf{Fixed dataset}. 
Facts might change in the world (and Wikidata). 
This can lead to a degraded real-world performance if a system relies exclusively on WebIE for evaluation when the dataset is not updated accordingly.

\section*{Acknowledgements}
We would like to thank Jens Lehmann for the helpful feedback on the paper draft, and Balkarn Hayre for helping with the MTurk experiments. 
We also thank the anonymous reviewers for their valuable comments that improved the paper.

\bibliography{custom}

\begin{thebibliography}{40}
\expandafter\ifx\csname natexlab\endcsname\relax\def\natexlab#1{#1}\fi

\bibitem[{Ayoola et~al.(2022)Ayoola, Tyagi, Fisher, Christodoulopoulos, and
  Pierleoni}]{ayoola-etal-2022-refined}
Tom Ayoola, Shubhi Tyagi, Joseph Fisher, Christos Christodoulopoulos, and
  Andrea Pierleoni. 2022.
\newblock \href {https://doi.org/10.18653/v1/2022.naacl-industry.24}
  {{R}e{F}in{ED}: An efficient zero-shot-capable approach to end-to-end entity
  linking}.
\newblock In \emph{Proceedings of the 2022 Conference of the North American
  Chapter of the Association for Computational Linguistics: Human Language
  Technologies: Industry Track}, pages 209--220, Hybrid: Seattle, Washington +
  Online. Association for Computational Linguistics.

\bibitem[{Bowman et~al.(2015)Bowman, Angeli, Potts, and
  Manning}]{bowman-etal-2015-large}
Samuel~R. Bowman, Gabor Angeli, Christopher Potts, and Christopher~D. Manning.
  2015.
\newblock \href {https://doi.org/10.18653/v1/D15-1075} {A large annotated
  corpus for learning natural language inference}.
\newblock In \emph{Proceedings of the 2015 Conference on Empirical Methods in
  Natural Language Processing}, pages 632--642, Lisbon, Portugal. Association
  for Computational Linguistics.

\bibitem[{Cao et~al.(2021)Cao, Izacard, Riedel, and
  Petroni}]{cao2021autoregressive}
Nicola~De Cao, Gautier Izacard, Sebastian Riedel, and Fabio Petroni. 2021.
\newblock \href {https://openreview.net/forum?id=5k8F6UU39V} {Autoregressive
  entity retrieval}.
\newblock In \emph{International Conference on Learning Representations}.

\bibitem[{Chaganty et~al.(2017)Chaganty, Paranjape, Liang, and
  Manning}]{chaganty-etal-2017-importance}
Arun Chaganty, Ashwin Paranjape, Percy Liang, and Christopher~D. Manning. 2017.
\newblock \href {https://doi.org/10.18653/v1/D17-1109} {Importance sampling for
  unbiased on-demand evaluation of knowledge base population}.
\newblock In \emph{Proceedings of the 2017 Conference on Empirical Methods in
  Natural Language Processing}, pages 1038--1048, Copenhagen, Denmark.
  Association for Computational Linguistics.

\bibitem[{Clark and Manning(2016)}]{clark-manning-2016-improving}
Kevin Clark and Christopher~D. Manning. 2016.
\newblock \href {https://doi.org/10.18653/v1/P16-1061} {Improving coreference
  resolution by learning entity-level distributed representations}.
\newblock In \emph{Proceedings of the 54th Annual Meeting of the Association
  for Computational Linguistics (Volume 1: Long Papers)}, pages 643--653,
  Berlin, Germany. Association for Computational Linguistics.

\bibitem[{Costa-juss{\`a} et~al.(2022)Costa-juss{\`a}, Cross, {\c{C}}elebi,
  Elbayad, Heafield, Heffernan, Kalbassi, Lam, Licht, Maillard
  et~al.}]{costa2022no}
Marta~R Costa-juss{\`a}, James Cross, Onur {\c{C}}elebi, Maha Elbayad, Kenneth
  Heafield, Kevin Heffernan, Elahe Kalbassi, Janice Lam, Daniel Licht, Jean
  Maillard, et~al. 2022.
\newblock No language left behind: Scaling human-centered machine translation.
\newblock \emph{arXiv preprint arXiv:2207.04672}.

\bibitem[{Eberts and Ulges(2020)}]{span2020}
Markus Eberts and Adrian Ulges. 2020.
\newblock Span-based joint entity and relation extraction with transformer
  pre-training.
\newblock \emph{ECAI}, page 2006–2013.

\bibitem[{Elsahar et~al.(2018)Elsahar, Vougiouklis, Remaci, Gravier, Hare,
  Laforest, and Simperl}]{elsahar-etal-2018-rex}
Hady Elsahar, Pavlos Vougiouklis, Arslen Remaci, Christophe Gravier, Jonathon
  Hare, Frederique Laforest, and Elena Simperl. 2018.
\newblock \href {https://aclanthology.org/L18-1544} {{T}-{RE}x: A large scale
  alignment of natural language with knowledge base triples}.
\newblock In \emph{Proceedings of the Eleventh International Conference on
  Language Resources and Evaluation ({LREC} 2018)}, Miyazaki, Japan. European
  Language Resources Association (ELRA).

\bibitem[{Ganitkevitch et~al.(2013)Ganitkevitch, Van~Durme, and
  Callison-Burch}]{ganitkevitch-etal-2013-ppdb}
Juri Ganitkevitch, Benjamin Van~Durme, and Chris Callison-Burch. 2013.
\newblock \href {https://aclanthology.org/N13-1092} {{PPDB}: The paraphrase
  database}.
\newblock In \emph{Proceedings of the 2013 Conference of the North {A}merican
  Chapter of the Association for Computational Linguistics: Human Language
  Technologies}, pages 758--764, Atlanta, Georgia. Association for
  Computational Linguistics.

\bibitem[{Gontier et~al.(2022)Gontier, Reddy, and Pal}]{gontier2022does}
Nicolas Gontier, Siva Reddy, and Christopher Pal. 2022.
\newblock \href {https://openreview.net/forum?id=9nhmKwLAWV} {Does entity
  abstraction help generative transformers reason?}
\newblock \emph{Transactions on Machine Learning Research}.

\bibitem[{Grycner and Weikum(2016)}]{grycner-weikum-2016-poly}
Adam Grycner and Gerhard Weikum. 2016.
\newblock \href {https://doi.org/10.18653/v1/D16-1236} {{POLY}: Mining
  relational paraphrases from multilingual sentences}.
\newblock In \emph{Proceedings of the 2016 Conference on Empirical Methods in
  Natural Language Processing}, pages 2183--2192, Austin, Texas. Association
  for Computational Linguistics.

\bibitem[{Han et~al.(2018)Han, Zhu, Yu, Wang, Yao, Liu, and
  Sun}]{han-etal-2018-fewrel}
Xu~Han, Hao Zhu, Pengfei Yu, Ziyun Wang, Yuan Yao, Zhiyuan Liu, and Maosong
  Sun. 2018.
\newblock \href {https://doi.org/10.18653/v1/D18-1514} {{F}ew{R}el: A
  large-scale supervised few-shot relation classification dataset with
  state-of-the-art evaluation}.
\newblock In \emph{Proceedings of the 2018 Conference on Empirical Methods in
  Natural Language Processing}, pages 4803--4809, Brussels, Belgium.
  Association for Computational Linguistics.

\bibitem[{Huguet~Cabot and
  Navigli(2021)}]{huguet-cabot-navigli-2021-rebel-relation}
Pere-Llu{\'\i}s Huguet~Cabot and Roberto Navigli. 2021.
\newblock \href {https://doi.org/10.18653/v1/2021.findings-emnlp.204} {{REBEL}:
  Relation extraction by end-to-end language generation}.
\newblock In \emph{Findings of the Association for Computational Linguistics:
  EMNLP 2021}, pages 2370--2381, Punta Cana, Dominican Republic. Association
  for Computational Linguistics.

\bibitem[{Ji et~al.(2020)Ji, Yu, Li, Ma, Wu, Tan, and Liu}]{ji-etal-2020-span}
Bin Ji, Jie Yu, Shasha Li, Jun Ma, Qingbo Wu, Yusong Tan, and Huijun Liu. 2020.
\newblock \href {https://doi.org/10.18653/v1/2020.coling-main.8} {Span-based
  joint entity and relation extraction with attention-based span-specific and
  contextual semantic representations}.
\newblock In \emph{Proceedings of the 28th International Conference on
  Computational Linguistics}, pages 88--99, Barcelona, Spain (Online).
  International Committee on Computational Linguistics.

\bibitem[{Johnson et~al.(2017)Johnson, Schuster, Le, Krikun, Wu, Chen, Thorat,
  Vi{\'e}gas, Wattenberg, Corrado, Hughes, and
  Dean}]{johnson-etal-2017-googles}
Melvin Johnson, Mike Schuster, Quoc~V. Le, Maxim Krikun, Yonghui Wu, Zhifeng
  Chen, Nikhil Thorat, Fernanda Vi{\'e}gas, Martin Wattenberg, Greg Corrado,
  Macduff Hughes, and Jeffrey Dean. 2017.
\newblock \href {https://doi.org/10.1162/tacl_a_00065} {{G}oogle{'}s
  multilingual neural machine translation system: Enabling zero-shot
  translation}.
\newblock \emph{Transactions of the Association for Computational Linguistics},
  5:339--351.

\bibitem[{Josifoski et~al.(2022)Josifoski, De~Cao, Peyrard, Petroni, and
  West}]{josifoski-etal-2022-genie}
Martin Josifoski, Nicola De~Cao, Maxime Peyrard, Fabio Petroni, and Robert
  West. 2022.
\newblock \href {https://doi.org/10.18653/v1/2022.naacl-main.342} {{G}en{IE}:
  Generative information extraction}.
\newblock In \emph{Proceedings of the 2022 Conference of the North American
  Chapter of the Association for Computational Linguistics: Human Language
  Technologies}, pages 4626--4643, Seattle, United States. Association for
  Computational Linguistics.

\bibitem[{Lewis et~al.(2020)Lewis, Liu, Goyal, Ghazvininejad, Mohamed, Levy,
  Stoyanov, and Zettlemoyer}]{lewis-etal-2020-bart}
Mike Lewis, Yinhan Liu, Naman Goyal, Marjan Ghazvininejad, Abdelrahman Mohamed,
  Omer Levy, Veselin Stoyanov, and Luke Zettlemoyer. 2020.
\newblock \href {https://doi.org/10.18653/v1/2020.acl-main.703} {{BART}:
  Denoising sequence-to-sequence pre-training for natural language generation,
  translation, and comprehension}.
\newblock In \emph{Proceedings of the 58th Annual Meeting of the Association
  for Computational Linguistics}, pages 7871--7880, Online. Association for
  Computational Linguistics.

\bibitem[{Mesquita et~al.(2019)Mesquita, Cannaviccio, Schmidek, Mirza, and
  Barbosa}]{mesquita-etal-2019-knowledgenet}
Filipe Mesquita, Matteo Cannaviccio, Jordan Schmidek, Paramita Mirza, and
  Denilson Barbosa. 2019.
\newblock \href {https://doi.org/10.18653/v1/D19-1069} {{K}nowledge{N}et: A
  benchmark dataset for knowledge base population}.
\newblock In \emph{Proceedings of the 2019 Conference on Empirical Methods in
  Natural Language Processing and the 9th International Joint Conference on
  Natural Language Processing (EMNLP-IJCNLP)}, pages 749--758, Hong Kong,
  China. Association for Computational Linguistics.

\bibitem[{Mintz et~al.(2009)Mintz, Bills, Snow, and
  Jurafsky}]{mintz-etal-2009-distant}
Mike Mintz, Steven Bills, Rion Snow, and Daniel Jurafsky. 2009.
\newblock \href {https://aclanthology.org/P09-1113} {Distant supervision for
  relation extraction without labeled data}.
\newblock In \emph{Proceedings of the Joint Conference of the 47th Annual
  Meeting of the {ACL} and the 4th International Joint Conference on Natural
  Language Processing of the {AFNLP}}, pages 1003--1011, Suntec, Singapore.
  Association for Computational Linguistics.

\bibitem[{Narayan et~al.(2022)Narayan, Sim{\~o}es, Zhao, Maynez, Das, Collins,
  and Lapata}]{narayan-etal-2022-well}
Shashi Narayan, Gon{\c{c}}alo Sim{\~o}es, Yao Zhao, Joshua Maynez, Dipanjan
  Das, Michael Collins, and Mirella Lapata. 2022.
\newblock \href {https://doi.org/10.18653/v1/2022.acl-long.94} {A well-composed
  text is half done! composition sampling for diverse conditional generation}.
\newblock In \emph{Proceedings of the 60th Annual Meeting of the Association
  for Computational Linguistics (Volume 1: Long Papers)}, pages 1319--1339,
  Dublin, Ireland. Association for Computational Linguistics.

\bibitem[{Narayan et~al.(2021)Narayan, Zhao, Maynez, Sim{\~o}es, Nikolaev, and
  McDonald}]{narayan-etal-2021-planning}
Shashi Narayan, Yao Zhao, Joshua Maynez, Gon{\c{c}}alo Sim{\~o}es, Vitaly
  Nikolaev, and Ryan McDonald. 2021.
\newblock \href {https://doi.org/10.1162/tacl_a_00438} {Planning with learned
  entity prompts for abstractive summarization}.
\newblock \emph{Transactions of the Association for Computational Linguistics},
  9:1475--1492.

\bibitem[{Nayak and Ng(2020)}]{Nayak_Ng_2020}
Tapas Nayak and Hwee~Tou Ng. 2020.
\newblock \href {https://doi.org/10.1609/aaai.v34i05.6374} {Effective modeling
  of encoder-decoder architecture for joint entity and relation extraction}.
\newblock \emph{Proceedings of the AAAI Conference on Artificial Intelligence},
  34(05):8528--8535.

\bibitem[{Ormandi et~al.(2021)Ormandi, Saleh, Winter, and
  Rao}]{ormandi2021webred}
Robert Ormandi, Mohammad Saleh, Erin Winter, and Vinay Rao. 2021.
\newblock Webred: Effective pretraining and finetuning for relation extraction
  on the web.
\newblock \emph{arXiv preprint arXiv:2102.09681}.

\bibitem[{Raffel et~al.(2020)Raffel, Shazeer, Roberts, Lee, Narang, Matena,
  Zhou, Li, and Liu}]{JMLR:v21:20-074}
Colin Raffel, Noam Shazeer, Adam Roberts, Katherine Lee, Sharan Narang, Michael
  Matena, Yanqi Zhou, Wei Li, and Peter~J. Liu. 2020.
\newblock \href {http://jmlr.org/papers/v21/20-074.html} {Exploring the limits
  of transfer learning with a unified text-to-text transformer}.
\newblock \emph{Journal of Machine Learning Research}, 21(140):1--67.

\bibitem[{Riedel et~al.(2010)Riedel, Yao, and McCallum}]{riedel2010}
Sebastian Riedel, Limin Yao, and Andrew McCallum. 2010.
\newblock Modeling relations and their mentions without labeled text.
\newblock In \emph{Machine Learning and Knowledge Discovery in Databases},
  pages 148--163, Berlin, Heidelberg. Springer Berlin Heidelberg.

\bibitem[{Rossiello et~al.(2023)Rossiello, Chowdhury, Mihindukulasooriya,
  Cornec, and Gliozzo}]{knowgl-aaai_2023_demo}
Gaetano Rossiello, Md. Faisal~Mahbub Chowdhury, Nandana Mihindukulasooriya,
  Owen Cornec, and Alfio Gliozzo. 2023.
\newblock Knowgl: Knowledge generation and linking from text.
\newblock In \emph{Proceedings of the AAAI Conference on Artificial
  Intelligence}.

\bibitem[{Seganti et~al.(2021)Seganti, Firl{\k{a}}g, Skowronska, Sat{\l}awa,
  and Andruszkiewicz}]{seganti-etal-2021-multilingual}
Alessandro Seganti, Klaudia Firl{\k{a}}g, Helena Skowronska, Micha{\l}
  Sat{\l}awa, and Piotr Andruszkiewicz. 2021.
\newblock \href {https://doi.org/10.18653/v1/2021.eacl-main.166} {Multilingual
  entity and relation extraction dataset and model}.
\newblock In \emph{Proceedings of the 16th Conference of the European Chapter
  of the Association for Computational Linguistics: Main Volume}, pages
  1946--1955, Online. Association for Computational Linguistics.

\bibitem[{Stoica et~al.(2021)Stoica, Platanios, and
  Poczos}]{Stoica_Platanios_Poczos_2021}
George Stoica, Emmanouil~Antonios Platanios, and Barnabas Poczos. 2021.
\newblock \href {https://doi.org/10.1609/aaai.v35i15.17631} {Re-tacred:
  Addressing shortcomings of the tacred dataset}.
\newblock \emph{Proceedings of the AAAI Conference on Artificial Intelligence},
  35(15):13843--13850.

\bibitem[{Sutskever et~al.(2014)Sutskever, Vinyals, and Le}]{NIPS2014_a14ac55a}
Ilya Sutskever, Oriol Vinyals, and Quoc~V Le. 2014.
\newblock \href
  {https://proceedings.neurips.cc/paper/2014/file/a14ac55a4f27472c5d894ec1c3c743d2-Paper.pdf}
  {Sequence to sequence learning with neural networks}.
\newblock In \emph{Advances in Neural Information Processing Systems},
  volume~27. Curran Associates, Inc.

\bibitem[{Tan et~al.(2022)Tan, Xu, Bing, Ng, and
  Aljunied}]{tan-etal-2022-revisiting}
Qingyu Tan, Lu~Xu, Lidong Bing, Hwee~Tou Ng, and Sharifah~Mahani Aljunied.
  2022.
\newblock \href {https://aclanthology.org/2022.emnlp-main.580} {Revisiting
  {D}oc{RED} - addressing the false negative problem in relation extraction}.
\newblock In \emph{Proceedings of the 2022 Conference on Empirical Methods in
  Natural Language Processing}, pages 8472--8487, Abu Dhabi, United Arab
  Emirates. Association for Computational Linguistics.

\bibitem[{Tang et~al.(2021)Tang, Tran, Li, Chen, Goyal, Chaudhary, Gu, and
  Fan}]{tang-etal-2021-multilingual}
Yuqing Tang, Chau Tran, Xian Li, Peng-Jen Chen, Naman Goyal, Vishrav Chaudhary,
  Jiatao Gu, and Angela Fan. 2021.
\newblock \href {https://doi.org/10.18653/v1/2021.findings-acl.304}
  {Multilingual translation from denoising pre-training}.
\newblock In \emph{Findings of the Association for Computational Linguistics:
  ACL-IJCNLP 2021}, pages 3450--3466, Online. Association for Computational
  Linguistics.

\bibitem[{Trisedya et~al.(2019)Trisedya, Weikum, Qi, and
  Zhang}]{trisedya-etal-2019-neural}
Bayu~Distiawan Trisedya, Gerhard Weikum, Jianzhong Qi, and Rui Zhang. 2019.
\newblock \href {https://doi.org/10.18653/v1/P19-1023} {Neural relation
  extraction for knowledge base enrichment}.
\newblock In \emph{Proceedings of the 57th Annual Meeting of the Association
  for Computational Linguistics}, pages 229--240, Florence, Italy. Association
  for Computational Linguistics.

\bibitem[{Vania et~al.(2022)Vania, Lee, and
  Pierleoni}]{vania-lee-and-andrea-pierleoni-2022-improving}
Clara Vania, Grace Lee, and Andrea Pierleoni. 2022.
\newblock \href {https://doi.org/10.18653/v1/2022.deeplo-1.2} {Improving
  distantly supervised document-level relation extraction through natural
  language inference}.
\newblock In \emph{Proceedings of the Third Workshop on Deep Learning for
  Low-Resource Natural Language Processing}, pages 14--20, Hybrid. Association
  for Computational Linguistics.

\bibitem[{Whitehouse et~al.(2023)Whitehouse, Weyde, and
  Madhyastha}]{whitehouse-etal-2023-towards}
Chenxi Whitehouse, Tillman Weyde, and Pranava Madhyastha. 2023.
\newblock \href {https://aclanthology.org/2023.findings-eacl.126} {Towards a
  unified model for generating answers and explanations in visual question
  answering}.
\newblock In \emph{Findings of the Association for Computational Linguistics:
  EACL 2023}, pages 1648--1660, Dubrovnik, Croatia. Association for
  Computational Linguistics.

\bibitem[{Whitehouse et~al.(2022)Whitehouse, Weyde, Madhyastha, and
  Komninos}]{whitehouse2022evaluation}
Chenxi Whitehouse, Tillman Weyde, Pranava Madhyastha, and Nikos Komninos. 2022.
\newblock Evaluation of fake news detection with knowledge-enhanced language
  models.
\newblock In \emph{Proceedings of the International AAAI Conference on Web and
  Social Media}, volume~16, pages 1425--1429.

\bibitem[{Williams et~al.(2018)Williams, Nangia, and
  Bowman}]{williams-etal-2018-broad}
Adina Williams, Nikita Nangia, and Samuel Bowman. 2018.
\newblock \href {https://doi.org/10.18653/v1/N18-1101} {A broad-coverage
  challenge corpus for sentence understanding through inference}.
\newblock In \emph{Proceedings of the 2018 Conference of the North {A}merican
  Chapter of the Association for Computational Linguistics: Human Language
  Technologies, Volume 1 (Long Papers)}, pages 1112--1122, New Orleans,
  Louisiana. Association for Computational Linguistics.

\bibitem[{Yamada et~al.(2020)Yamada, Asai, Shindo, Takeda, and
  Matsumoto}]{yamada-etal-2020-luke}
Ikuya Yamada, Akari Asai, Hiroyuki Shindo, Hideaki Takeda, and Yuji Matsumoto.
  2020.
\newblock \href {https://doi.org/10.18653/v1/2020.emnlp-main.523} {{LUKE}: Deep
  contextualized entity representations with entity-aware self-attention}.
\newblock In \emph{Proceedings of the 2020 Conference on Empirical Methods in
  Natural Language Processing (EMNLP)}, pages 6442--6454, Online. Association
  for Computational Linguistics.

\bibitem[{Yao et~al.(2021)Yao, Du, Lin, Li, Liu, Zhou, and
  Sun}]{yao-etal-2021-codred}
Yuan Yao, Jiaju Du, Yankai Lin, Peng Li, Zhiyuan Liu, Jie Zhou, and Maosong
  Sun. 2021.
\newblock \href {https://doi.org/10.18653/v1/2021.emnlp-main.366} {{C}od{RED}:
  A cross-document relation extraction dataset for acquiring knowledge in the
  wild}.
\newblock In \emph{Proceedings of the 2021 Conference on Empirical Methods in
  Natural Language Processing}, pages 4452--4472, Online and Punta Cana,
  Dominican Republic. Association for Computational Linguistics.

\bibitem[{Yao et~al.(2019)Yao, Ye, Li, Han, Lin, Liu, Liu, Huang, Zhou, and
  Sun}]{yao-etal-2019-docred}
Yuan Yao, Deming Ye, Peng Li, Xu~Han, Yankai Lin, Zhenghao Liu, Zhiyuan Liu,
  Lixin Huang, Jie Zhou, and Maosong Sun. 2019.
\newblock \href {https://doi.org/10.18653/v1/P19-1074} {{D}oc{RED}: A
  large-scale document-level relation extraction dataset}.
\newblock In \emph{Proceedings of the 57th Annual Meeting of the Association
  for Computational Linguistics}, pages 764--777, Florence, Italy. Association
  for Computational Linguistics.

\bibitem[{Zhang et~al.(2017)Zhang, Zhong, Chen, Angeli, and
  Manning}]{zhang-etal-2017-position}
Yuhao Zhang, Victor Zhong, Danqi Chen, Gabor Angeli, and Christopher~D.
  Manning. 2017.
\newblock \href {https://doi.org/10.18653/v1/D17-1004} {Position-aware
  attention and supervised data improve slot filling}.
\newblock In \emph{Proceedings of the 2017 Conference on Empirical Methods in
  Natural Language Processing}, pages 35--45, Copenhagen, Denmark. Association
  for Computational Linguistics.

\end{thebibliography}
\bibliographystyle{acl_natbib}

\appendix








\begin{table*}[!ht]
\centering
\scalebox{0.7}{
\addtolength{\tabcolsep}{-3.4pt}
\begin{tabular}{ll|cccrcccrcccccccccc}
\toprule
&{\multirow{2}{*}{\sc Model}}  &\multicolumn{4}{c} {\sc WebIE (all test)}   &\multicolumn{4}{c}  {\sc WebIE (anno. test)}& \multicolumn{3}{c} {\sc REBEL}   & \multicolumn{3}{c} {\sc Wiki-NRE}
\\
&& \textit{Precision} &\textit{ Recall} & \textit{F1}  & \textit{Acc.-Neg.}& \textit{Precision} &\textit{ Recall} & \textit{F1}  & \textit{Acc.-Neg.} & \textit{Precision} &\textit{ Recall} & \textit{F1 }& \textit{Precision} & \textit{Recall} &\textit{ F1 }  \\
\cmidrule(lr){1-2} 
\cmidrule(lr){3-6} \cmidrule(lr){7-10} \cmidrule(lr){11-13}  \cmidrule(lr){14-16} 
\multirow{7}{*}{\sc {\rotatebox[origin=c]{90}{\small{\textsc{Unconstrained}}}}} &
\sc Bart\textsubscript{rand} (r)  
&  \cellcolor[HTML]{DDEBF7}10.83
& \cellcolor[HTML]{DDEBF7}16.00
& \cellcolor[HTML]{DDEBF7}12.92
& \cellcolor[HTML]{DDEBF7}0.00
&  \cellcolor[HTML]{DDEBF7}10.70		
&  \cellcolor[HTML]{DDEBF7}13.26
& \cellcolor[HTML]{DDEBF7}11.84  
& \cellcolor[HTML]{DDEBF7}0.00
& 64.34	&67.90&	66.07 		
&  \cellcolor[HTML]{DDEBF7}15.83
& \cellcolor[HTML]{DDEBF7}52.09 
& \cellcolor[HTML]{DDEBF7}24.28\\
&\sc Bart\textsubscript{plm} (r) 
& \cellcolor[HTML]{DDEBF7}17.58
& \cellcolor[HTML]{DDEBF7}34.20
&  \cellcolor[HTML]{DDEBF7}23.23
&	 \cellcolor[HTML]{DDEBF7}2.28
&  \cellcolor[HTML]{DDEBF7}17.95		
&  \cellcolor[HTML]{DDEBF7}30.02
& \cellcolor[HTML]{DDEBF7}22.47  
& \cellcolor[HTML]{DDEBF7}1.97
& 63.83	&76.66&	69.66
&  \cellcolor[HTML]{DDEBF7}18.34		
& \cellcolor[HTML]{DDEBF7}65.04
& \cellcolor[HTML]{DDEBF7}28.62
\\
\cmidrule(lr){2-16}
&\sc Bart\textsubscript{rand} (w)
& 55.06&	54.90&	54.98&89.67
&51.64&44.46&47.78&94.74
&  \cellcolor[HTML]{DDEBF7}22.45	
& \cellcolor[HTML]{DDEBF7}20.42 
& \cellcolor[HTML]{DDEBF7}21.39
&  \cellcolor[HTML]{DDEBF7}10.95
& \cellcolor[HTML]{DDEBF7}31.49
& \cellcolor[HTML]{DDEBF7}16.25\\
&\sc Bart\textsubscript{plm} (w)   

& 54.81 & 70.29 & 61.59 & 87.59 
& 53.40 & 62.36 & 57.53 & 93.58
&\cellcolor[HTML]{DDEBF7} 28.05 
&\cellcolor[HTML]{DDEBF7} 37.28 
&\cellcolor[HTML]{DDEBF7} 32.01
&\cellcolor[HTML]{DDEBF7} 15.55 
&\cellcolor[HTML]{DDEBF7} 60.45 
&\cellcolor[HTML]{DDEBF7} 24.73
\\
\cmidrule(lr){2-16}
&\sc Bart\textsubscript{rand} (r+w)   
&51.34 & 61.22 & 55.85 & 86.80
&49.64 &51.62 &50.61& 93.15
&64.38 &69.57 & 66.87
&  \cellcolor[HTML]{DDEBF7}17.68	
& \cellcolor[HTML]{DDEBF7}65.96
& \cellcolor[HTML]{DDEBF7}27.89
\\
&\sc Bart\textsubscript{plm} (r+w)
& 53.04	&75.29&	62.23 & 76.66
&  53.18&	68.41&	59.84& 82.96
 &63.49 &75.30	&	68.89
&  \cellcolor[HTML]{DDEBF7}18.93
& \cellcolor[HTML]{DDEBF7}73.52
& \cellcolor[HTML]{DDEBF7}30.11
\\

\midrule
\multirow{7}{*}{\sc {\rotatebox[origin=c]{90}{\small{\textsc{Constraint Trie}}}}} &
\sc Bart\textsubscript{rand} (r) 		
&  \cellcolor[HTML]{DDEBF7}11.93
&  \cellcolor[HTML]{DDEBF7}18.91
& \cellcolor[HTML]{DDEBF7}14.63 
& \cellcolor[HTML]{DDEBF7}0.00
&  \cellcolor[HTML]{DDEBF7}11.82		
&  \cellcolor[HTML]{DDEBF7}15.63
& \cellcolor[HTML]{DDEBF7}13.46  
& \cellcolor[HTML]{DDEBF7}0.00
& 66.89&70.37&	68.58 
&\cellcolor[HTML]{DDEBF7}27.61
&\cellcolor[HTML]{DDEBF7}66.73
&\cellcolor[HTML]{DDEBF7}39.06\\
&\sc Bart\textsubscript{plm} (r) 		
&  \cellcolor[HTML]{DDEBF7}15.24
&  \cellcolor[HTML]{DDEBF7}39.30 
& \cellcolor[HTML]{DDEBF7}21.96
&  \cellcolor[HTML]{DDEBF7}0.00
&  \cellcolor[HTML]{DDEBF7}15.98		
&  \cellcolor[HTML]{DDEBF7}34.92
& \cellcolor[HTML]{DDEBF7}21.93 
& \cellcolor[HTML]{DDEBF7}0.00
&66.28&	76.78 &	71.14
&\cellcolor[HTML]{DDEBF7}25.39
&	\cellcolor[HTML]{DDEBF7}77.45
&	\cellcolor[HTML]{DDEBF7}38.24
\\
\cmidrule(lr){2-16}
&\sc Bart\textsubscript{rand} (w)
&  55.47 &	57.25&	56.35&90.07
&  52.95&	46.60&	49.57&95.04
&\cellcolor[HTML]{DDEBF7}27.47
&\cellcolor[HTML]{DDEBF7}23.13
&	\cellcolor[HTML]{DDEBF7}25.12
&\cellcolor[HTML]{DDEBF7}18.98
&	\cellcolor[HTML]{DDEBF7}43.75
&\cellcolor[HTML]{DDEBF7}26.48
\\
&\sc Bart\textsubscript{plm} (w) 

& 57.92 & 74.19 & 64.91 & 87.99
& 57.00 & 65.91 & 61.13 & 94.18
&\cellcolor[HTML]{DDEBF7}35.81 
&\cellcolor[HTML]{DDEBF7}43.00 
&\cellcolor[HTML]{DDEBF7}39.08
&\cellcolor[HTML]{DDEBF7}24.30 
&\cellcolor[HTML]{DDEBF7}78.01 
&\cellcolor[HTML]{DDEBF7}37.06
\\
\cmidrule(lr){2-16}
&\sc Bart\textsubscript{rand} (r+w)  
&52.79 &64.15 &57.92 & 87.45
&51.89& 54.28 & 53.06  &  93.71
&66.87 &72.24 &69.45
& \cellcolor[HTML]{DDEBF7}29.02
& \cellcolor[HTML]{DDEBF7}82.35
&\cellcolor[HTML]{DDEBF7}42.91
\\
&\sc Bart\textsubscript{plm} (r+w)
&54.63&78.43&64.40 & 76.43
&55.22&71.25&62.22&82.59
&66.42 &78.29 & 71.87
& \cellcolor[HTML]{DDEBF7}29.25 
& \cellcolor[HTML]{DDEBF7}86.38
& \cellcolor[HTML]{DDEBF7}43.70
\\
\bottomrule
\end{tabular}
}
\caption{Additional results using beam search with and without constraint Trie for each dataset.
Results in blue shades are zero-shot performance.
}
\label{tab:main-full}
\end{table*}

\begin{table*}[!ht]
\centering
\vspace{0.4cm}
\scalebox{0.75}{
\addtolength{\tabcolsep}{-2pt}
\begin{tabular}{l|rrrrc|rrrrc}
\toprule
{\multirow{2}{*}{\sc Language}}
 &\multicolumn{5}{c|} { EN as Source Language in mBART Tokenizer}  &\multicolumn{5}{c} {XX as Source Language in mBART Tokenizer}
\\

&\textit{Precision} &\textit{ Recall} & \textit{F1} & \textit{Empty-Pos.\%} 
& \textit{Accuracy-Neg.} 
&\textit{Precision} &\textit{ Recall} & \textit{F1} & \textit{Empty-Pos.\%}  & \textit{Accuracy-Neg.} 

\\
\cmidrule(lr){1-1} 
\cmidrule(lr){2-4} \cmidrule(lr){5-5} \cmidrule(lr){6-6} \cmidrule(lr){7-9}  
\cmidrule(lr){10-10} 
\cmidrule(lr){11-11} 

  \sc French 
  &43.27 &36.13 &39.38 &11.89 &96.19
&41.29 & 37.73 & 39.43 & 8.56 & 94.87
\\
 \sc Spanish  
&41.93 & 34.63 & 37.93 &12.34 & 96.74
& 40.47 & 36.57 & 38.42 & 8.56 & 95.82
\\
\sc Portuguese
&41.17 &32.37 &36.24 &14.07 &96.91
&13.81 &1.77 & 3.14 & 86.33 & 98.21
\\
 \sc Hindi 
  &4.28 & 1.62 & 2.35 &67.38& 98.64
& 3.69 & 1.69 & 2.31 & 60.62 & 98.43
\\
\bottomrule
\end{tabular}
}
\caption{Comparison of the zero-shot performance on m\textsc{WebIE} with mBART when specifying the source language (XX) and keeping the default setting as the source language (EN).
Results are with standard beam search (without the constraint Trie).
}
\label{tab:src-lang}
\end{table*}

\begin{table*}[ht]
    \centering
    \scalebox{0.8}{
    \setlength{\tabcolsep}{.5em}
    \begin{tabulary}{1.0\textwidth}{lp{15em}L}
    \toprule
    \textbf{Example Id} & \textbf{Sentence} & \textbf{ReFinED Output} \\
    \midrule
    21464177
    & On Thursday, British campaigning group the Environmental Investigation Agency accused Italy of trying to sabotage efforts to reform the EU ETS. 
    & [["Thursday", None, "DATE"], ["British", Entity(wikidata\_entity\_id=Q145, wikipedia\_entity\_title=United Kingdom), "GPE"], ["Environmental Investigation Agency", Entity(wikidata\_entity\_id=Q1345905, wikipedia\_entity\_title=Environmental Investigation Agency), "ORG"], ["Italy", Entity(wikidata\_entity\_id=Q38, wikipedia\_entity\_title=Italy), "ORG"], ["EU", Entity(wikidata\_entity\_id=Q458, wikipedia\_entity\_title=European Union), "ORG"], ["ETS", Entity(wikidata\_entity\_id=Q899383, wikipedia\_entity\_title=ETSI), "ORG"]] \\
    \midrule
    \addlinespace[.15cm]
    1274217 
    & It culminates in the decade-long debate ending in 1913 to turn the Hetch Hetchy valley in Yosemite National Park into a reservoir for San Francisco. 
    & [['decade-long', None, 'DATE'], ['1913', Entity(parsed\_string=[timepoint: ["1913"]]), 'DATE'], ['Hetch Hetchy', Entity(wikidata\_entity\_id=Q1616130, wikipedia\_entity\_title=Hetch Hetchy), 'GPE'], ['Yosemite National Park', Entity(wikidata\_entity\_id=Q180402, wikipedia\_entity\_title=Yosemite National Park), 'FAC'], ['San Francisco', Entity(wikidata\_entity\_id=Q62, wikipedia\_entity\_title=San Francisco), 'GPE']] \\
    \bottomrule
    \end{tabulary}
    }
    \caption{ReFinED outputs on \textsc{WebIE} validation examples.}
    \label{tab:refined-examples}
\end{table*}

\section{Additional Results}
\label{sec:main-full}
We show the full results in \autoref{tab:main-full} for \textsc{Bart\textsubscript{rand}} and \textsc{Bart\textsubscript{plm}} trained on REBEL and \webie, using both beam search with and without constraint Trie decoding.

We show in \autoref{tab:src-lang} the results for non-English languages for m\webie\ when specifying the source language and using the default (English) for the mBART tokenizer. 
These results are from beam search without constraint Trie.
We can see that specifying the source language mostly harms the performance (except French), especially for Portuguese.
We hypothesise that due to the model being trained solely on English as the source token, mBART may have difficulty handling other languages. 

\begin{figure*}[ht]
\centering
\vspace{0.4cm}
    \includegraphics[width=0.95\linewidth]{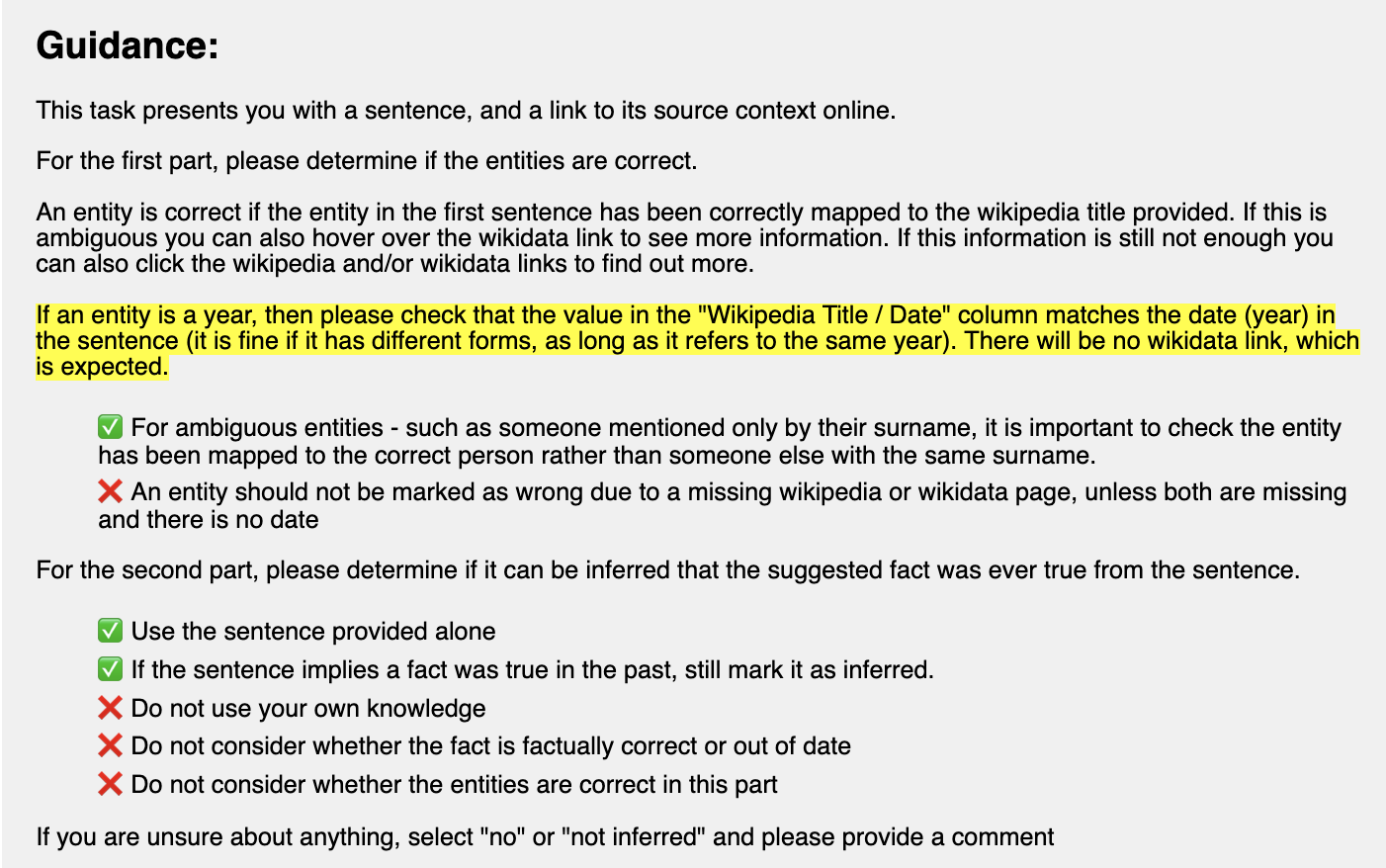}
    \caption{MTurk HIT guidance entity and relation labelling. }
\label{fig:guidance}
\end{figure*}

\begin{figure*}[th]
\centering
    \includegraphics[width=0.95\linewidth]{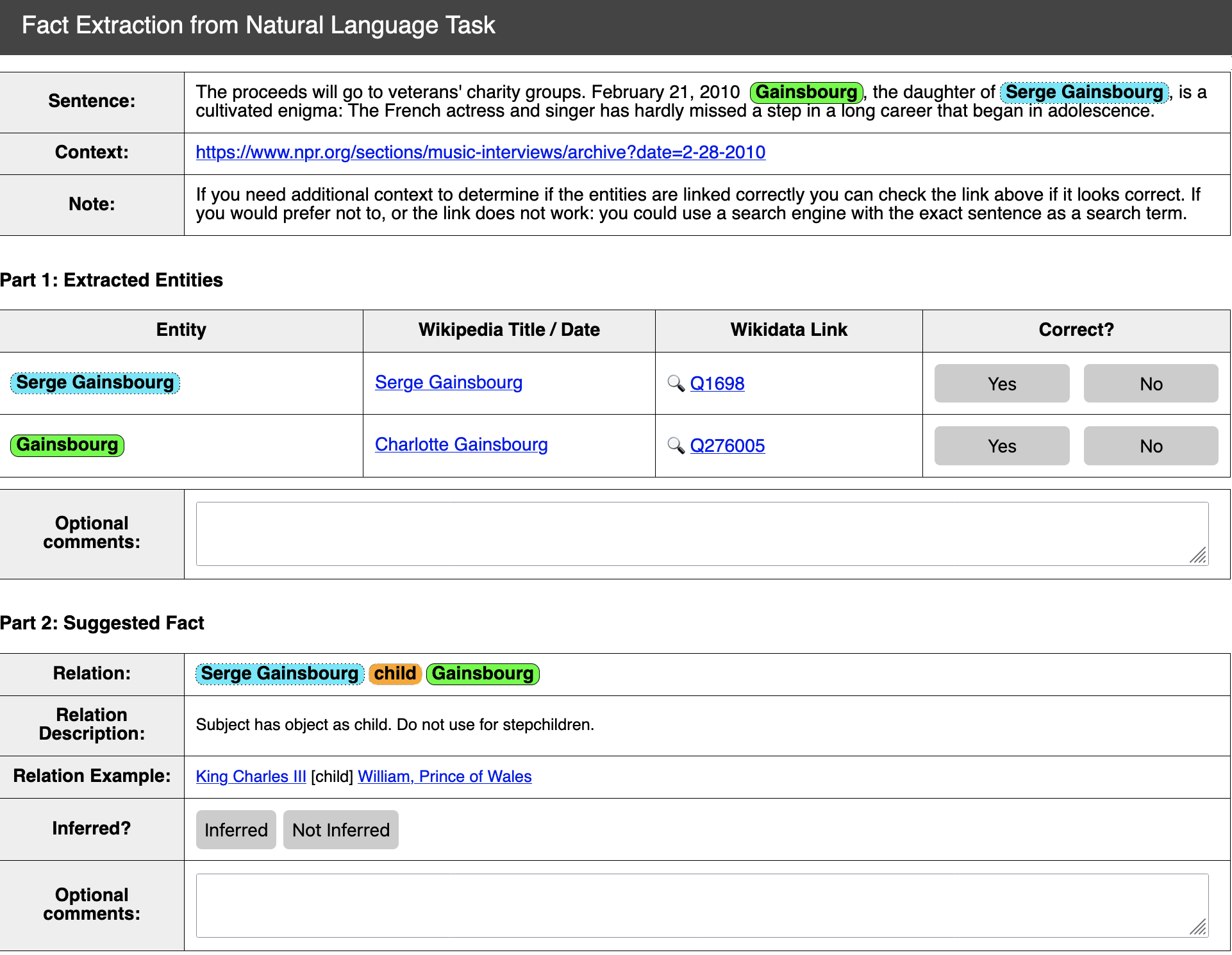}
    \caption{MTurk HIT user interface for entity and relation labelling. }
\label{fig:webie_mturk1}
\end{figure*}

\begin{figure*}[ht]
\centering
    \includegraphics[width=0.95\linewidth]{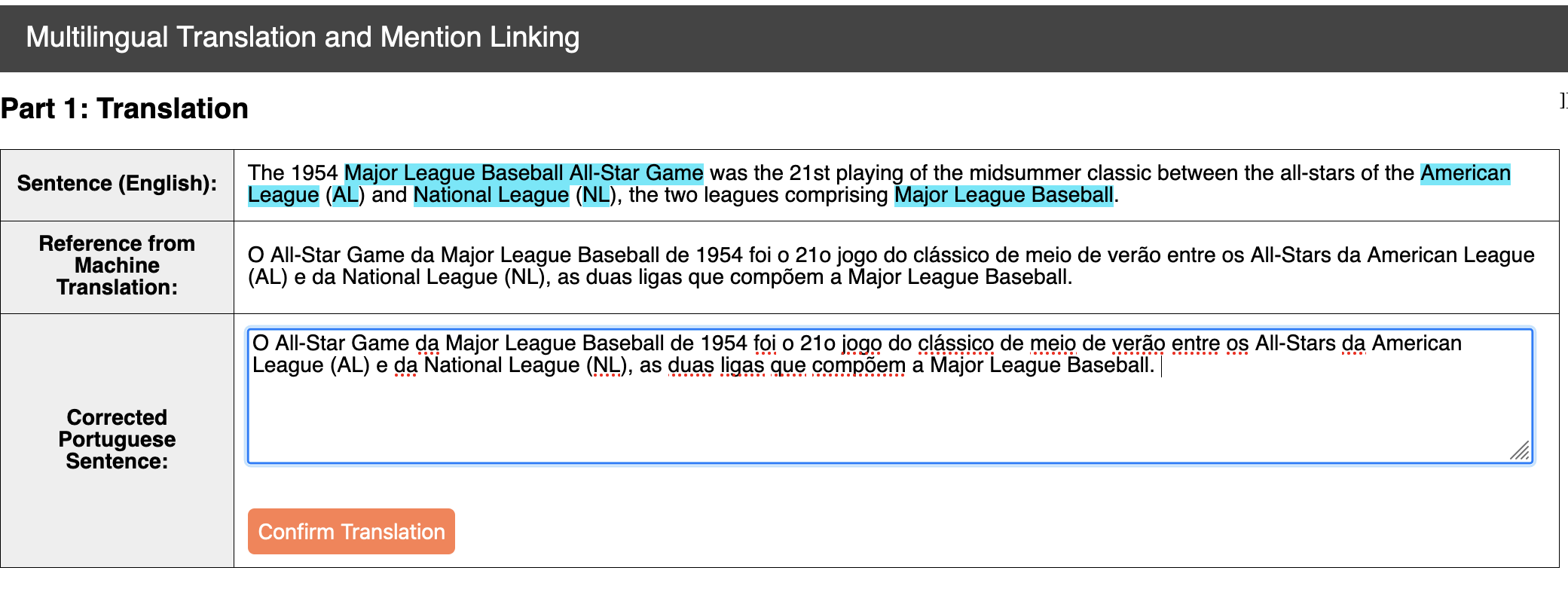}
    \caption{MTurk HIT user interface for correcting the machine-translated text. }
\label{fig:webie_mturk2}
\end{figure*}

\begin{figure*}[ht]
\centering
    \includegraphics[width=0.95\linewidth]{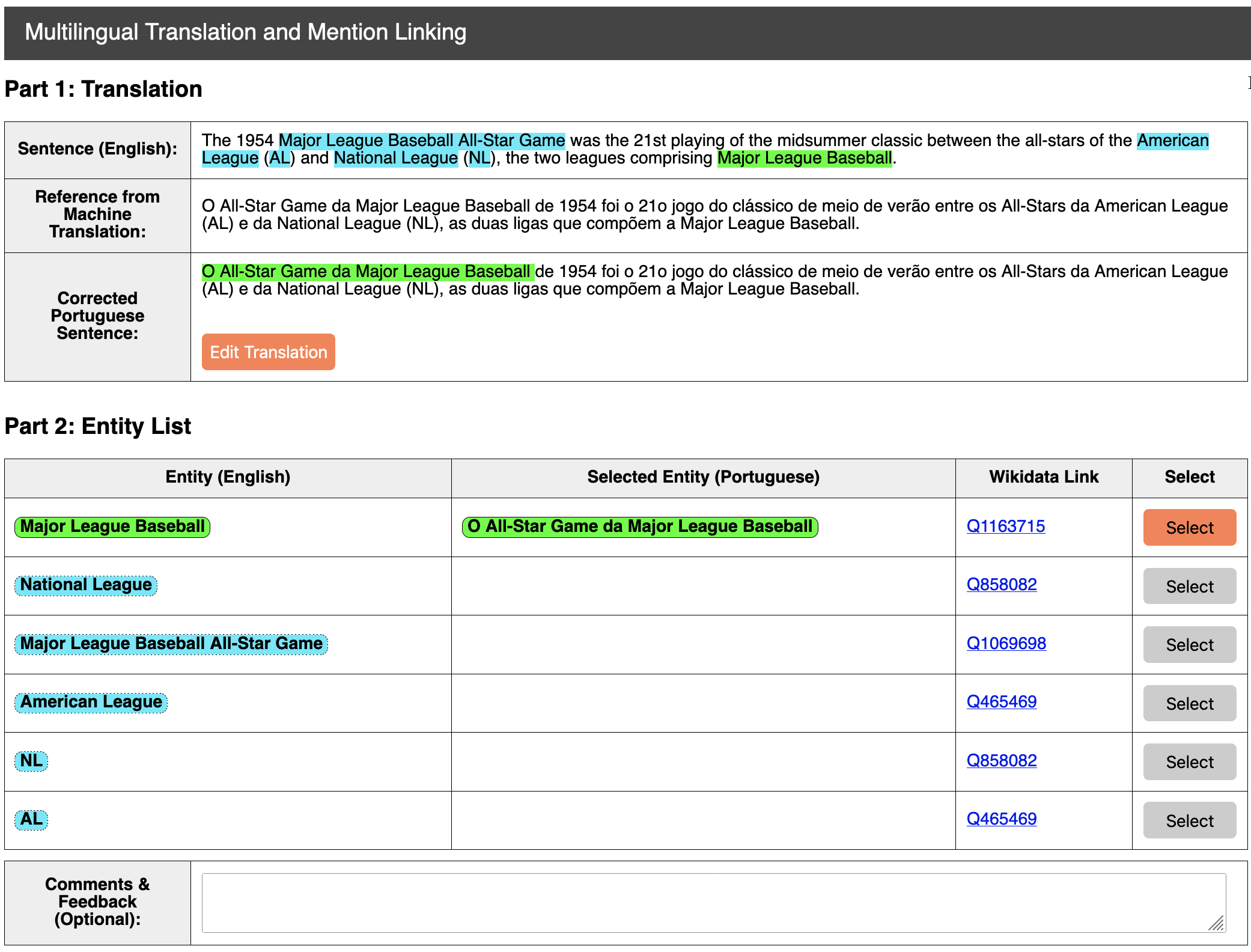}
    \caption{MTurk HIT user interface for entity labelling in the target language. }
\label{fig:webie_mturk3}
\end{figure*}

\section{Examples of ReFinED Output}
\label{sec:ReFinED}

We show examples of the sentences processed by ReFinED in \autoref{tab:refined-examples}. For each input sentence, ReFinED identifies the set of entities in that sentence, and outputs mention span, Wikidata id, and Wikipedia title for each entity. For our experiments, we use the \texttt{wikipedia\_model\_with\_numbers} model with \texttt{wikipedia} entity set.

\section{MTurk Annotation Details}
\label{sec:mturk}
In this section, we describe the detailed settings for annotating (m)\webie with MTurk.

\subsection{\textsc{WebIE}}
The first annotation task (HIT) is to verify the correctness of the triples automatically created from the DS approach and filtered by the NLI model. The guidance and the interface are shown in \autoref{fig:guidance} and \autoref{fig:webie_mturk1}, respectively. 

In each HIT, we provide a sentence with its entities highlighted (head entity in blue and tail entity in green) and the URL of the web page which the sentence is extracted from.
For the first EL annotation job, we provide both links to the Wikipedia and Wikidata pages.
Annotators are asked to choose if the highlighted spans are linked correctly to the KB.
Next, the annotators are asked to verify if a relation (highlighted in orange) can be inferred from the sentence.
We provide the description of the relation and an example use case to facilitate the annotation.

Each triple is annotated by three workers, and we pay \$0.2 per HIT.
We hire MTurk workers with Masters Qualification and set additional requirements including (i) having done 2,000 HITs and (ii) having a job approval rate $\geq$99\%.

\subsection{m\textsc{WebIE}}
\label{sec:anno-mwebie}

\autoref{fig:webie_mturk2} and \autoref{fig:webie_mturk3} illustrates the interface for correcting machine-translated sentence and identifying corresponding entities in them.
As it is challenging to find qualified crowd workers for the translation task\footnote{Preliminary results where we include the USA for the m\webie\ annotation task indicate that MTurk workers with limited or no knowledge of the target language (or English) still accept the job, despite our specific requirement for proficiency in both English and the target language.}, we set the geographical regions for each language to the countries where the language is one of the official languages.
We find that only India and countries in America have an adequate number of MTurk workers, which highly restricts the options for our target languages.
In the end, the countries we set for the target languages are as follows: 
Portuguese: AO, BR, CV, ST, GW, GQ, MZ; 
Spanish: ES, MX, CO, PE, CL; CA for French, and IN for Hindi\footnote{For the mapping between country codes and countries, please refer to \url{https://docs.aws.amazon.com/AWSMechTurk/latest/AWSMturkAPI/ApiReference_LocaleDataStructureArticle.html}}.
It was also necessary to remove the Masters Qualification requirement for MTurk workers (except Hindi) to find adequate annotators.
We then conduct pilot annotations, where we deliberately introduce errors in the reference machine translation to verify if the workers under our requirement settings are able to correct them.

We provide the English sentence paired with the original machine-translated sentence for the actual HIT.
The English sentence is highlighted with its entity spans, and we instruct the workers to correct the translation while ensuring that the entities are correctly translated.
After confirming the translation, workers are then asked to highlight the corresponding entities in the target language (in green).
For negative sentences without entity spans, the longest noun phrases were highlighted instead to prevent workers from simply copying the reference translations.
We pay \$0.35 per HIT for positive sentences and \$0.25 for negative sentences (since most sentences in negative examples have only one highlighted entity/noun phrase and it is considered an easier task).

Two MTurk workers are asked for the translation task, and an additional worker was asked to select the better translation, for which \$0.10 per HIT was paid.

\section{Domains in \textsc{WebIE}}
\label{sec:domains}

The 200 URL domains included in \textsc{WebIE} are shown in \autoref{tab:domains}.

\section{Relations in the Annotated Set}
\label{sec:200_rel}

\autoref{tab:relation4} shows the details of the 200 relations that are covered in the human-annotated set of \webie.

\begin{table*}[ht]
\centering
\scalebox{0.75}{
\addtolength{\tabcolsep}{0pt}
\begin{tabular} {l|l|l|l}
\toprule
www.nytimes.com
&www.latimes.com
&www.theguardian.com
&www.businessinsider.com \\www.forbes.com
&www.chicagotribune.com
&www.foxnews.com
&www.aljazeera.com\\www.dailymail.co.uk
&www.express.co.uk
&www.cnet.com
&www.telegraph.co.uk\\www.rt.com
&www.zdnet.com
&www.foxbusiness.com
&www.reuters.com\\www.ibtimes.co.uk
&www.washingtonpost.com
&www.si.com
&www.bbc.com\\news.bbc.co.uk
&nypost.com
&www.marketwired.com
&www.baltimoresun.com\\www.npr.org
&www.fool.com
&www.bbc.co.uk
&mashable.com\\www.cnbc.com
&www.hindustantimes.com
&www.csmonitor.com
&www.yahoo.com\\www.thesun.co.uk
&www.nydailynews.com
&www.dailystar.co.uk
&www.kickstarter.com\\uk.reuters.com
&www.inquisitr.com
&www.straitstimes.com
&www.cbsnews.com\\deadline.com
&www.androidheadlines.com
&www.wired.com
&www.bustle.com\\www.pcworld.com
&www.fastcompany.com
&www.firstpost.com
&www.entrepreneur.com\\www.breitbart.com
&techcrunch.com
&www.nme.com
&www.ndtv.com\\finance.yahoo.com
&www.lonelyplanet.com
&www.ign.com
&www.barnesandnoble.com\\www.usatoday.com
&www.timeout.com
&apnews.com
&www.thisisinsider.com\\metro.co.uk
&gizmodo.com
&www.sacbee.com
&economictimes.indiatimes.com\\www.buzzfeed.com
&www.miamiherald.com
&www.espn.com
&www.washingtontimes.com\\www.pbs.org
&thenextweb.com
&www.aol.com
&
timesofindia.indiatimes.com\\www.cbc.ca
&kotaku.com
&www.irishtimes.com
&www.military.com\\www.startribune.com
&www.deccanherald.com
&www.techradar.com
&www.thestar.com\\www.techrepublic.com
&slate.com
&www.pcmag.com
&www.hollywoodreporter.com\\www.marketwatch.com
&www.slideshare.net
&www.etonline.com
&in.reuters.com\\variety.com
&www.sfgate.com
&indianexpress.com
&www.abc.net.au\\theconversation.com
&www.eurekalert.org
&mic.com
&www.blogtalkradio.com\\www.thenation.com
&www.prnewswire.com
&www.barrons.com
&www.apnews.com\\www.newsmax.com
&www.theatlantic.com
&www.huffpost.com
&patents.google.com\\www.eventbrite.com
&link.springer.com
&www.ncbi.nlm.nih.gov
&www.prweb.com\\www.deviantart.com
&www.instructables.com
&www.booking.com
&www.etsy.com\\sites.google.com
&www.agreatertown.com
&lists.w3.org
&disneyparksmomspanel.disney.go.com\\homestars.com
&www.reference.com
&www.city-data.com
&app-wiringdiagram.herokuapp.com\\www.adweek.com
&docs.microsoft.com
&fineartamerica.com
&www.insiderpages.com\\lists.debian.org
&premium.wpmudev.org
&www.librarything.com
&mail-archives.apache.org\\scholars.duke.edu
&www.glassdoor.com
&www.shutterstock.com
&myemail.constantcontact.com\\www.eventbrite.co.uk
&archives.lib.state.ma.us
&www.gsmarena.com
&www.audible.com\\www.hotels.com
&www.statista.com
&www.alibaba.com
&lists.gnu.org\\ipfs.io
&www.socialbakers.com
&www.weddingwire.com
&rd.springer.com\\appadvice.com
&www.complex.com
&zapier.com
&www.foodnetwork.com\\www.kijiji.ca
&www.salon.com
&www.semanticscholar.org
&hubpages.com\\www.scribd.com
&www.cinemablend.com
&w3techs.com
&www.urbandictionary.com\\www.salespider.com
&www.angieslist.com
&stackoverflow.com
&www.dictionary.com\\www.zocdoc.com
&wordpress.org
&www.pcgamer.com
&www.chamberofcommerce.com\\www.worldcat.org
&s3.amazonaws.com
&www.tweaktown.com
&chroniclingamerica.loc.gov\\www.agoda.com
&www.showmelocal.com
&www.refinery29.com
&www.businessinsider.com.au

\\www.healthgrades.com
&store.cdbaby.com

&oppositelock.kinja.com
&www.bedbathandbeyond.com\\www.radionz.co.nz
&www.ebay.com
&downloads.zdnet.com
&www.stitcher.com\\www.thestreet.com
&github.com
&www.youtube.com
&www.oreilly.com\\itunes.apple.com
&medium.com
&www.tripadvisor.com
&www.imdb.com\\forums.newtek.com
&forums.macrumors.com
&answers.sap.com
&forum.duolingo.com\\community.esri.com
&en.wikipedia.org
&en.m.wikipedia.org
&encyclopedia2.thefreedictionary.com\\simple.wikipedia.org
&www.encyclopedia.com
&www.britannica.com
&www.questia.com\\
\bottomrule
\end{tabular}
}
\caption{URL domains of the sentences included in \webie.
}
\label{tab:domains}
\end{table*}

\clearpage

\begin{table*}[!ht]
\centering
\scalebox{0.7}{
\addtolength{\tabcolsep}{0pt}
\begin{tabular} {p{1cm}|p{1cm}|p{3.8cm}|p{14.4cm}}
\toprule
\multicolumn{1}{l|}{\textsc{Count}} & \textsc{Pid} & \textsc{Relation}                                            & \textsc{Description}    \\
\midrule
1359& P17  & country          & sovereign state of this item (not to be used for human beings)        \\
910 & P131 & located in the administrative territorial entity  & the item is located on the territory of the following   administrative entity. Use P276 for specifying locations that are   non-administrative places and for items about events. Use P1382 if the item   falls only partially into the administrative entity.   \\
776 & P530 & diplomatic relation         & diplomatic relations of the country  \\
684 & P47  & shares border with          & countries or administrative subdivisions, of equal level, that   this item borders, either by land or water. A single common point is enough.      \\
655 & P27    & country of citizenship      & the object is a country that recognizes the subject as its   citizen  \\
588 & P161   & cast member      & actor in the subject production .use "character   role" (P453) and/or "name of the character role" (P4633) as   qualifiers, use "voice actor" (P725) for voice-only role \\
580 & P577   & publication date & date or point in time when a work was first published or   released   \\
546 & P527   & has part(s)      & part of this subject      \\
480 & P54    & member of sports team       & sports teams or clubs that the subject represents or   represented    \\
438 & P800   & notable work     & notable scientific, artistic or literary work, or other work   of significance among subject's works   \\
437 & P463   & member of        & organization, club or musical group to which the subject belongs. Do not   use for membership in ethnic or social groups, nor for holding a political   position, such as a member of parliament (use P39 for that). \\
430 & P108   & employer         & person or organization for which the subject works or worked          \\
426 & P127   & owned by         & owner of the subject      \\
400 & P361   & part of          & object of which the subject is a part (if this subject is   already part of object A which is a part of object B, then please only make   the subject part of object A)  \\
378 & P1830  & owner of         & entities owned by the subject        \\
370 & P102   & member of political party   & the political party of which a person is or has been a member   or otherwise affiliated     \\
364 & P150   & contains the administrative territorial entity    & (list of) direct subdivisions of an administrative territorial   entity          \\
359 & P749   & parent organization         & parent organization of an organization, opposite of subsidiaries      \\
340 & P178   & developer        & organization or person that developed the item  \\
314 & P159   & headquarters location       & city, where an organization's headquarters is or has been situated. Use   (P276) qualifier for specific building  \\
310 & P57    & director         & director(s) of film, TV-series, stageplay, video game or   similar    \\
299 & P118   & league& league in which team or player plays or has played in      \\
297 & P1376  & capital of       & country, state, department, canton or other administrative   division of which the municipality is the governmental seat     \\
296 & P449   & original broadcaster        & network(s) or service(s) that originally broadcast a radio or   television program          \\
293 & P36    & capital          & seat of government of a country, province, state or other type   of administrative territorial entity  \\
285 & P2936  & language used    & language widely used (spoken or written) in this place or at   this event        \\
280 & P355   & has subsidiary   & subsidiary of a company or organization; generally a fully   owned separate corporation     \\
279 & P175   & performer        & actor, musician, band or other performer associated with this   role or musical work        \\
267 & P166   & award received   & award or recognition received by a person, organization or   creative work       \\
267 & P569   & date of birth    & date on which the subject was born   \\
262 & P641   & sport & sport that the subject participates or participated in or is   associated with   \\
258 & P26    & spouse& the subject has the object as their spouse (husband, wife, partner,   etc.). Use "unmarried partner" (P451)) for non-married companions \\
247 & P571   & inception        & time when an entity begins to exist; for date of official opening use   P1619    \\
241 & P176   & manufacturer     & manufacturer or producer of this product        \\
234 & P40    & child & subject has object as child. Do not use for stepchildren   \\
233 & P170   & creator          & maker of this creative work or other object (where no more   specific property exists)      \\
227 & P3373  & sibling          & the subject and the object have at   least one common parent (brother, sister, etc. including half-siblings); use   "relative" (P1038) for siblings-in-law (brother-in-law,   sister-in-law, etc.) and step-siblings (step-brothers, step-sisters, etc.)         \\
227 & P50    & author& main creator(s) of a written work (use on works, not humans);   use P2093 when Wikidata item is unknown or does not exist    \\
226 & P570   & date of death    & date on which the subject died       \\
224 & P276   & location         & location of the object, structure or event. In the case of an   administrative entity as containing item use P131. For statistical entities   use P8138. In the case of a geographic entity use P706. Use P7153 for locations   associated with the object.      \\
204 & P674   & characters       & characters which appear in this item (like plays, operas,   operettas, books, comics, films, TV series, video games)         \\
203 & P1412  & languages spoken, written or signed    & language(s) that a person or a people speaks, writes or signs,   including the native language(s)      \\
201 & P1441  & present in work  & this (fictional or fictionalized) entity or person appears in   that work as part of the narration (use P2860 for works citing other works, :P361/P1433 for works   being part of other works, P1343 for entities described in non-fictional   accounts)         \\
201 & P945   & allegiance       & country (or other power) that the person or group serves   \\
197 & P58    & screenwriter     & person(s) who wrote the script for subject item \\
197 & P37    & official language& language designated as official by this item    \\
193 & P137   & operator         & person, profession, or organization that operates the   equipment, facility, or service     \\
193 & P162   & producer         & person(s) who produced the film, musical work, theatrical   production, etc. (for film, this does not include executive producers,   associate producers, etc.)  \\
185 & P1411  & nominated for    & award nomination received by a person, organisation or creative work   \\
\bottomrule
\end{tabular}
\label{tab:relation}
}
\end{table*}

\clearpage

\begin{table*}[!ht]
\centering
\scalebox{0.7}{
\addtolength{\tabcolsep}{0pt}
\begin{tabular} {p{1cm}|p{1cm}|p{3.8cm}|p{14.4cm}}
\toprule
\multicolumn{1}{l|}{\textsc{Count}} & \textsc{Pid} & \textsc{Relation}                                            & \textsc{Description}    \\
\midrule

184 & P1056  & product or material produced& material or product produced by a government agency, business,   industry, facility, or process        \\
183 & P35    & head of state    & official with the highest formal authority in a country/state         \\
180 & P206   & located in or next to body of water    & body of water on or next to which a place is located       \\
180 & P1001  & applies to jurisdiction     & the item (institution, law, public office, public register...)   or statement belongs to or has power over or applies to the value (a   territorial jurisdiction: a country, state, municipality, ...)    \\
180 & P144   & based on         & the work(s) used as the basis for subject item  \\
177 & P156   & followed by      & immediately following item in a series of which the subject is   a part, preferably use as qualifier of P179      \\
176 & P112   & founded by       & founder or co-founder of this organization, religion or place         \\
174 & P155   & follows          & immediately prior item in a series of which the subject is a   part, preferably use as qualifier of P179          \\
171 & P488   & chairperson      & presiding member of an organization, group or body         \\
169 & P279   & subclass of      & this item is a subclass (subset) of   that item; all instances of these items are instances of those items;   different from P31 (instance of), e.g.: K2 is an instance of mountain;   volcano is a subclass of mountain (and an instance of volcanic landform). \\
169 & P169   & chief executive officer     & highest-ranking corporate officer appointed as the CEO within   an organization  \\
168 & P86    & composer         & person(s) who wrote the music {[}for lyricist, use "lyrics by"   (P676)          \\
164 & P140   & religion or worldview       & religion of a person, organization or religious building, or   associated with this subject \\
163 & P750   & distributed by   & distributor of a creative work; distributor for a record   label; news agency; film distributor        \\
161 & P974   & tributary        & watercourse that flows into an other one (for lake inflows use   P200)\\
159 & P6087  & coach of sports team        & sports club or team for which this person is or was on-field   manager or coach  \\
157 & P197   & adjacent station & the stations next to this station, sharing the same line(s)\\
156 & P1344  & participant in   & event in which a person or organization was/is a participant          \\
155 & P272   & production company          & company that produced this film, audio or performing arts work        \\
154 & P461   & opposite of      & item that is the opposite of this item          \\
152 & P1365  & replaces         & person, state or item replaced. Use "structure   replaces" (P1398) for structures.          \\
152 & P277   & programmed in    & the programming language(s) in which the software is developed        \\
151 & P19    & place of birth   & most specific known (e.g. city instead of country, or hospital   instead of city) birth location of a person, animal or fictional character        \\
150 & P1366  & replaced by      & other person or item which continues the item by replacing it in its   role. Use P156 ("followed by") if the item is not replaced nor   identical, but adds to the series (e.g. books in a series).       \\
148 & P585   & point in time    & time and date something took place, existed or a statement was   true \\
148 & P710   & participant      & person, group of people or organization (object) that actively   takes/took part in an event or process (subject).  Preferably qualify   with "object has role" (P3831)). Use P1923   for participants that are teams.          \\
147 & P466   & occupant         & person or organization occupying property       \\
144 & P7047  & enemy of         & opponent character or group of this fictive character or group        \\
143 & P580   & start time       & time a time period starts \\
138 & P403   & mouth of the watercourse    & the body of water to which the watercourse drains          \\
135 & P400   & platform         & platform for which a work was developed or released, or the   specific platform version of a software product     \\
134 & P1327  & partner in business or sport& professional collaborator \\
134 & P22    & father& male parent of the subject. Not stepfather      \\
134 & P414   & stock exchange   & exchange on which this company is traded        \\
133 & P306   & operating system & operating system (OS) on which a software works or the OS   installed on hardware\\
129 & P1346  & winner& winner of a competition or similar event, NOT to be used for   awards (instead use "award received" on awardee's item, possibly   qualified with "for work" or for wars or battles  \\
128 & P1889  & different from   & item that is different from another item, with which it may be   confused        \\
128 & P4969  & derivative work  & new work of art (film, book, software, etc.) derived from   major part of this work         \\
127 & P31    & instance of      & that class of which this subject is a particular example and   member; different from 'subclass of'; for example: K2 is an instance of   mountain; volcano is a subclass of mountain (and an instance of volcanic   landform)   \\
127 & P30    & continent        & continent of which the subject is a part        \\
124 & P397   & parent astronomical body    & major astronomical body the item belongs to     \\
122 & P607   & conflict         & battles, wars or other military engagements in which the   person or item participated      \\
120 & P2789  & connects with    & item with which the item is physically connected\\
120 & P1038  & relative         & family member (qualify with "type of kinship"; for   direct family member please use specific property)\\
119 & P1891  & signatory        & person, country, or organization that has signed an official   document          \\
118 & P1029  & crew member(s)   & person(s) that participated operating or serving aboard this   vehicle\\
115 & P937   & work location    & location where persons or organisations were actively   participating in employment, business or other work       \\
114 & P495   & country of origin& country of origin of this item (creative work, food, phrase,   product, etc.)    \\
113 & P1557  & manifestation of & inherent and characteristic embodiment of a given concept  \\
113 & P6     & head of government          & head of the executive power of this town, city, municipality,   state, country, or other governmental body        \\
\bottomrule
\end{tabular}
}
\label{tab:relation2}
\end{table*}

\clearpage

\begin{table*}[!ht]
\centering
\scalebox{0.7}{
\addtolength{\tabcolsep}{0pt}
\begin{tabular} {p{1cm}|p{1cm}|p{3.8cm}|p{14.4cm}}
\toprule
\multicolumn{1}{l|}{\textsc{Count}} & \textsc{Pid} & \textsc{Relation}                                            & \textsc{Description}    \\
\midrule

112 & P451   & unmarried partner& someone with whom the person is in a relationship without being married.   Use "spouse" (P26) for married couples \\
112 & P725   & voice actor      & performer of a spoken role in a creative work such as   animation, video game, radio drama, or dubbing over       \\
112 & P123   & publisher        & organization or person responsible for publishing books,   periodicals, printed music, podcasts, games or software\\
111 & P264   & record label     & brand and trademark associated with the marketing of subject   music recordings and music videos       \\
111 & P737   & influenced by    & this person, idea, etc. is informed by that other person,   idea, etc., e.g. “Heidegger was influenced by Aristotle”         \\
110 & P706   & located in/on physical feature         & located on the specified (geo)physical feature. Should not be   used when the value is only political/administrative (P131) or a   mountain range (P4552).    \\
109 & P3095  & practiced by     & type of agents that study this subject or work in this   profession   \\
108 & P1716  & brand & commercial brand associated with the item       \\
106 & P115   & home venue       & home stadium or venue of a sports team or applicable   performing arts organization         \\
106 & P3461  & designated as terrorist by  & country or organization that has officially designated a given group as a   terrorist organization     \\
105 & P136   & genre & creative work's genre or an artist's field of work. Use main subject P921   to relate creative works to their topic          \\
105 & P69    & educated at      & educational institution attended by subject     \\
105 & P1532  & country for sport& country a person or a team represents when playing a sport \\
104 & P172   & ethnic group     & subject's ethnicity (consensus is that a VERY high standard of   proof is needed for this field to be used. In general this means 1) the   subject claims it themselves, or 2) it is widely agreed on by scholars, or 3)   is fictional and portrayed as such)   \\
103 & P205   & basin country    & country that have drainage to/from or border the body of water        \\
103 & P20    & place of death   & most specific known (e.g. city instead of country, or hospital   instead of city) death location of a person, animal or fictional character        \\
102 & P1923  & participating team          & like 'Participant' (P710) but for   teams. For an event like a cycle race or a football match you can use this   property to list the teams and P710 to list the   individuals (with 'member of sports team' P54 as a qualifier for the individuals)  \\
101 & P398   & child astronomical body     & minor body that belongs to the item  \\
100 & P179   & part of the series          & series which contains the subject    \\
100 & P450   & astronaut mission& space mission that the subject is or has been a member of (do   not include future missions)\\
99  & P25    & mother& female parent of the subject. Not stepmother    \\
99  & P8345  & media franchise  & this creative work belongs to this media franchise         \\
98  & P582   & end time         & time a time period ends   \\
97  & P2341  & indigenous to    & place or ethnic group where a language, art genre, cultural   tradition or expression, cooking style or food, or biological species or   variety is found (or was originally found) \\
95  & P194   & legislative body & legislative body governing this entity; political institution   with elected representatives, such as a parliament/legislature or council          \\
95  & P840   & narrative location          & the narrative of the work is set in this location          \\
95  & P287   & designed by      & person(s) or organization which designed the object        \\
94  & P103   & native language  & language or languages a person has learned from early   childhood     \\
94  & P7959  & historic county  & traditional, geographical division of Great Britain and   Ireland     \\
93  & P113   & airline hub      & airport that serves as a hub for an airline     \\
93  & P121   & item operated    & equipment, installation or service operated by the subject \\
91  & P6886  & writing language & language in which the writer has written their work        \\
91  & P6379  & has works in the collection & collection that has works of this person or organisation (use   archive location P485 for the archives)\\
90  & P126   & maintained by    & person or organization in charge of keeping the subject (for   instance an infrastructure) in functioning order   \\
90  & P647   & drafted by       & which team the player was drafted by \\
90  & P551   & residence        & the place where the person is or has been, resident        \\
90  & P3342  & significant person          & person linked to the item in any possible way   \\
89  & P1431  & executive producer          & executive producer of a movie or TV show        \\
88  & P1416  & affiliation      & organization that a person or organization is affiliated with   (not necessarily member of or employed by)        \\
88  & P138   & named after      & entity or event that inspired the subject's name, or namesake   (in at least one language)  \\
87  & P2031  & work period (start)         & start of period during which a person or group flourished in   their professional activity  \\
87  & P241   & military branch  & branch to which this military unit, award, office, or person   belongs, e.g. Royal Navy     \\
87  & P2541  & operating area   & geographic area or jurisdiction an organisation or industry   operates in, serves, or has responsibility for      \\
86  & P676   & lyrics by        & author of song lyrics     \\
86  & P1191  & date of first performance   & date a work was first debuted, performed or live-broadcasted          \\
85  & P190   & twinned administrative body & twin towns, sister cities, twinned municipalities and other   localities that have a partnership or cooperative agreement, either legally   or informally acknowledged by their governments    \\
84  & P598   & commander of   & for persons who are notable as commanding officers, the units   they commanded   \\
84  & P84    & architect        & person or architectural firm responsible for designing this   building\\
83  & P1336  & territory claimed by        & administrative divisions that claim control of a given area\\
82  & P199   & business division& organizational divisions of this organization (which are not   independent legal entities)  \\
\bottomrule
\end{tabular}
}
\label{tab:relation3}
\end{table*}

\clearpage

\begin{table*}[!ht]
\centering
\scalebox{0.7}{
\addtolength{\tabcolsep}{0pt}
\begin{tabular} {p{1cm}|p{1cm}|p{3.8cm}|p{14.4cm}}
\toprule
\multicolumn{1}{l|}{\textsc{Count}} & \textsc{Pid} & \textsc{Relation}                                            & \textsc{Description}    \\
\midrule

82  & P915   & filming location & actual place where this scene/film was shot. For the setting, use   "narrative location" (P840)        \\
82  & P371   & presenter        & main role in presenting a radio or television program or a   performing arts show\\
80  & P740   & location of formation       & location where a group or organization was formed          \\
79  & P2512  & series spin-off  & series' spin-offs         \\
79  & P1382  & partially coincident with   & object that partially overlaps with the subject in its   instances, parts, or members       \\
79  & P291   & place of publication        & geographical place of publication of the edition (use 1st   edition when referring to works)\\
78  & P39    & position held    & subject currently or formerly holds the object position or   public office       \\
78  & P1535  & used by          & item or concept that makes use of the subject (use   sub-properties when appropriate)       \\
78  & P1027  & conferred by     & person or organization who grants an award, certification,   grant, or role      \\
78  & P210   & party chief representative  & chief representative of a party in an institution or an   administrative unit \\
76  & P1269  & facet of         & topic of which this item is an aspect, item that offers a   broader perspective on the same topic      \\
75  & P4913  & dialect of       & language of which an item with this property is a dialect. Use   in addition to "subclass of" (P279) if a languoid is also   considered a dialect. \\
75  & P1619  & date of official opening    & date or point in time an event, museum, theater etc.   officially opened         \\
75  & P208   & executive body   & branch of government for the daily administration of the   territorial entity    \\
75  & P376   & located on astronomical body& astronomical body on which features or places are situated \\
74  & P931   & place served by transport hub          & territorial entity or entities served by this transport hub   (airport, train station, etc.)\\
74  & P793   & significant event& significant or notable events associated with the subject  \\
73  & P8138  & located in the statistical territorial entity     & statistical territorial entity in which a place is located or   is part of. If a municipality or county is split into or part of several   regions: add several values   \\
73  & P2032  & work period (end)& end of period during which a person or group flourished in   their professional activity    \\
73  & P3842  & located in the present-day administrative territorial entity & the item was located in the territory of this present-day   administrative unit; however the two did not at any point coexist in time   \\
71  & P664   & organizer        & person or institution organizing an event       \\
71  & P6872  & has written for  & publication an author has contributed to        \\
71  & P747   & has edition or translation  & link to an edition of this item      \\
71  & P1951  & investor         & individual or organization which invests money in the item for   the purpose of obtaining financial return on their investment          \\
69  & P576   & dissolved, abolished or demolished date& point in time at which the subject (organisation, building)   ceased to exist;  see "date of official closure" (P3999) for closing   a facility, "service retirement" (P730) for retiring   equipment,  "discontinued date" (P2669) for stopping a product       \\
69  & P101   & field of work    & specialization of a person or organization      \\
69  & P1408  & licensed to broadcast to    & place that a radio/TV station is licensed/required to   broadcast to  \\
69  & P832   & public holiday   & official public holiday that occurs in this place in its   honor, usually a non-working day \\
68  & P61    & discoverer or inventor      & subject who discovered, first described, invented, or   developed this discovery or invention          \\
68  & P38    & currency         & currency used by item     \\
68  & P1142  & political ideology          & political ideology of an organization or person or of a work (such as a   newspaper)        \\
67  & P1435  & heritage designation        & heritage designation of a cultural or natural site         \\
67  & P119   & place of burial  & location of grave, resting place, place of ash-scattering,   etc. (e.g., town/city or cemetery) for a person or animal. There may be   several places: e.g., re-burials, parts of body buried separately. \\
67  & P286   & head coach       & on-field manager or head coach of a sports club (not to be confused with   a general manager P505, which is not a coaching position) or person     \\
67  & P797   & authority        & entity having executive power on given entity   \\
66  & P364   & original language of film or TV show   & language in which a film or a performance work was originally created.   Deprecated for written works and songs; use P407 ("language of work or   name") instead.        \\
65  & P413   & position played on team / speciality   & position or specialism of a player on a team    \\
65  & P1304  & central bank     & country's central bank    \\
65  & P921   & main subject     & primary topic of a work   \\
65  & P3975  & secretary general& leader of a political or international organization, sometimes   below the chairperson      \\
64  & P1037  & director / manager          & person who manages any kind of group \\
64  & P407   & language of work or name    & language associated with this creative work (such as books,   shows, songs, broadcasts or websites) or a name (for persons use "native   language" P103 and   "languages spoken, written or signed"  P1412\\
63  & P177   & crosses          & obstacle (body of water, road, railway...) which this bridge   crosses over or this tunnel goes under  \\
63  & P3033  & package management system   & package management system used to publish the software     \\
62  & P1877  & after a work by  & artist whose work strongly inspired/ was copied in this item          \\
60  & P98    & editor& person who checks and correct a work (such as a book,   newspaper, academic journal, etc.) to comply with a rules of certain genre      \\
58  & P729   & service entry    & date or point in time on which a piece or class of equipment   entered operational service  \\
57  & P3301  & broadcast by     & channel, network, website or service that broadcast this item   over radio, TV or the Internet         \\
55  & P726   & candidate        & person or party that is an option for an office in this   election    \\
53  & P4884  & court & specific court a legal case is/was heard/decided in        \\
53  & P669   & located on street& street, road, or square, where the item is located.       
\\
\bottomrule
\end{tabular}
}
\caption{Count, PID (Wikidata ID), Relations and Descriptions of the top 200 relations in the annotated \textsc{WebIE}.
}
\label{tab:relation4}
\end{table*}

\clearpage

\end{document}